\documentclass[pdflatex,sn-basic]{sn-jnl}

\jyear{2023}%

\theoremstyle{thmstyleone}%
\theoremstyle{thmstyletwo}%

\theoremstyle{thmstylethree}%

\normalbaroutside

\usepackage{multirow}
\usepackage{lscape}
\usepackage{booktabs}
\usepackage{xcolor}
\usepackage{subcaption}
\usepackage{pdflscape} 
\usepackage{afterpage}
\usepackage{tikz}
\usepackage{rotating}

\usetikzlibrary{decorations.pathmorphing}

\definecolor{cadmiumgreen}{rgb}{0.0, 0.42, 0.24}
\definecolor{cadmiumred}{rgb}{0.89, 0.0, 0.13}

\newcommand{\new}[1]{\textcolor{black}{#1}}

\normalbaroutside 
\usepackage[acronym, nonumberlist]{glossaries}
\glsdisablehyper 

\makenoidxglossaries

\newacronym{gdpr}{GDPR}{General Data Protection Regulation}
\newacronym{ai}{AI}{Artificial Intelligence}
\newacronym{auc}{AUC}{Area Under the ROC curve}
\newacronym{automl}{AutoML}{Automated Machine Learning}
\newacronym{hp}{HP}{Hyperparameter}
\newacronym{ml}{ML}{Machine Learning}
\newacronym{dt}{DT}{Decision Tree}
\newacronym{ann}{ANN}{Artificial Neural Network}
\newacronym{svm}{SVM}{Support Vector Machine}
\newacronym{cart}{CART}{Classification and Regression Tree}
\newacronym{nbtree}{NBTree}{Na{\"i}ve-Bayes Tree}
\newacronym{lmt}{LMT}{Logistic Model Tree}
\newacronym{ctree}{CTree}{Conditional Inference Trees}
\newacronym{ga}{GA}{Genetic Algorithm}
\newacronym{pso}{PSO}{Particle Swarm Optimization}
\newacronym{eda}{EDA}{Estimation of Distribution Algorithm}
\newacronym{rs}{RS}{Random Search}
\newacronym{smbo}{SMBO}{Sequential Model-based Optimization}
\newacronym{irace}{Irace}{Iterated F-race}
\newacronym{rf}{RF}{Random Forest}
\newacronym{ps}{PS}{Pattern Search}
\newacronym{vtj48}{VTJ48}{Visual Tuning J48}
\newacronym{uci}{UCI}{University of California Irvine}
\newacronym{ss}{SS}{Scatter Search}
\newacronym{gp}{GP}{Gaussian Process}
\newacronym{gs}{GS}{Grid Search}
\newacronym{mtl}{MtL}{Meta-learning}
\newacronym{cash}{CASH}{Combined Algorithm Selection and Hyper-parameter Optimization}
\newacronym{cv}{CV}{Cross-validation}
\newacronym{openml}{OpenML}{Open Machine Learning}
\newacronym{cd}{CD}{Critical Difference}
\newacronym{ber}{BER}{Balanced Error Rate}
\newacronym{bac}{BAC}{Balanced per class Accuracy}
\newacronym{dl}{DL}{Deep Learning}
\newacronym{sh}{SH}{Shrinking Hypercube}
\newacronym{pd}{PD}{Parametric Density}     
\newacronym{chaid}{CHAID}{Chi-square Automatic Interaction Detection}
\newacronym{bohb}{BOHB}{Bayesian Optimization with HyperBand}
\newacronym{rep}{REP}{Reduced Error Pruning}
\newacronym{knn}{kNN}{k-Nearest Neighbors}
\newacronym{nb}{NB}{Na\"ive Bayes}
\newacronym{lr}{LR}{Logistic Regression}
\newacronym{shap}{SHAP}{SHapley Additive exPlanations}
\newacronym{xai}{XAI}{eXplanable Artificial Intelligence}

\begin{document}

\title
[Better Trees]
{Better Trees: An empirical study on hyperparameter tuning of \new{classification} decision tree induction algorithms}


\author[1]{\fnm{Rafael} \sur{Gomes Mantovani}}\email{rafaelmantovani@utfpr.edu.br}

\author[2,3]{\fnm{Tom\'a\v{s}} \sur{Horv\'ath}}\email{tomas.horvath@inf.elte.hu}

\author[4]{\fnm{André} \sur{L. D. Rossi}}\email{{andre.rossi@unesp.br}}

\author[5]{\fnm{Ricardo} \sur{Cerri}}\email{cerri@ufscar.br}

\author[6]{\fnm{Sylvio} \sur{Barbon Junior}}\email{{sylvio.barbonjunior@units.it}}

\author[7]{\fnm{Joaquin} \sur{Vanschoren}}\email{{j.vanschoren@tue.nl}}

\author[8]{\fnm{Andr{\'e}} \sur{C. P. L. F. {de Carvalho}}}\email{{andre@icmc.usp.br}}

\affil[1]{\orgname{Federal Technology University, Paran\'a, Campus of Apucarana, \orgaddress{\city{Apucarana}, \state{PR}, \country{Brazil}}}}

\affil[2]{\orgname{\new{Pavol Jozef \v{S}af\'arik University, Faculty of Science, Institute of Computer Science}}, \orgaddress{\new{\city{Ko\v{s}ice}, \country{Slovakia}}}}

\affil[3]{\orgname{ELTE E\"{o}tv\"{o}s Lor\'and University, Faculty of Informatics}, \orgaddress{\city{Budapest}, \country{Hungary}}}

\affil[4]{\orgname{São Paulo State University (Unesp), Campus of Itapeva}, \orgaddress{\city{Itapeva}, \state{SP}, \country{Brazil}}}

\affil[5]{\orgname{\new{Department of Computer Science}, Federal University of São Carlos, \orgaddress{\city{São Carlos}, \state{SP}, \country{Brazil}}}}

\affil[6]{\orgname{University of Trieste (UniTS)}, \orgaddress{\city{Trieste}, \country{Italy}}}

\affil[7]{\orgname{Eindhoven University of Technology (TU/e)}, \orgaddress{\city{Eindhoven}, \country{The Netherlands}}}

\affil[8]{\orgname{Institute of Mathematics and Computer Sciences (ICMC), University of S\~ao Paulo (USP), \orgaddress{\city{São Carlos}, \state{SP}, \country{Brazil}}}}


\abstract{
Machine learning algorithms often contain many hyperparameters whose values affect the predictive performance of the induced models in intricate ways. Due to the high number of possibilities for these hyperparameter configurations and their complex interactions, it is common to use optimization techniques to find settings that lead to high predictive performance.
However, insights into efficiently exploring this vast space of configurations and dealing with the trade-off between predictive and runtime performance remain challenging. 
Furthermore, there are cases where the default hyperparameters fit the suitable configuration.
Additionally, for many reasons, including model validation and attendance to new legislation, there is an increasing interest in interpretable models, such as those created by the~\acrfull{dt} induction algorithms.
This paper provides a comprehensive approach for investigating the effects of hyperparameter tuning for the two DT induction algorithms most often used, \acrshort{cart} and C4.5.
DT induction algorithms present high predictive performance and interpretable classification models, though many hyperparameters need to be adjusted. 
Experiments were carried out with different tuning strategies to induce models and to evaluate hyperparameters' relevance
using 94 classification datasets from OpenML. 
The experimental results point out that different hyperparameter profiles for the tuning of each algorithm provide statistically significant improvements in most of the datasets for \acrshort{cart}, but only in one-third for C4.5.
Although different algorithms may present different tuning scenarios, the tuning techniques generally required few evaluations to find accurate solutions. Furthermore, the best technique for all the algorithms was the \acrshort{irace}. 
Finally, we found out that tuning a specific small subset of hyperparameters is a good alternative for achieving optimal predictive performance.
}


\maketitle


\section{Introduction}
\label{sec:intro}

\begin{figure}[!ht]
    \centering
    \includegraphics
    [width=0.9\textwidth]
    {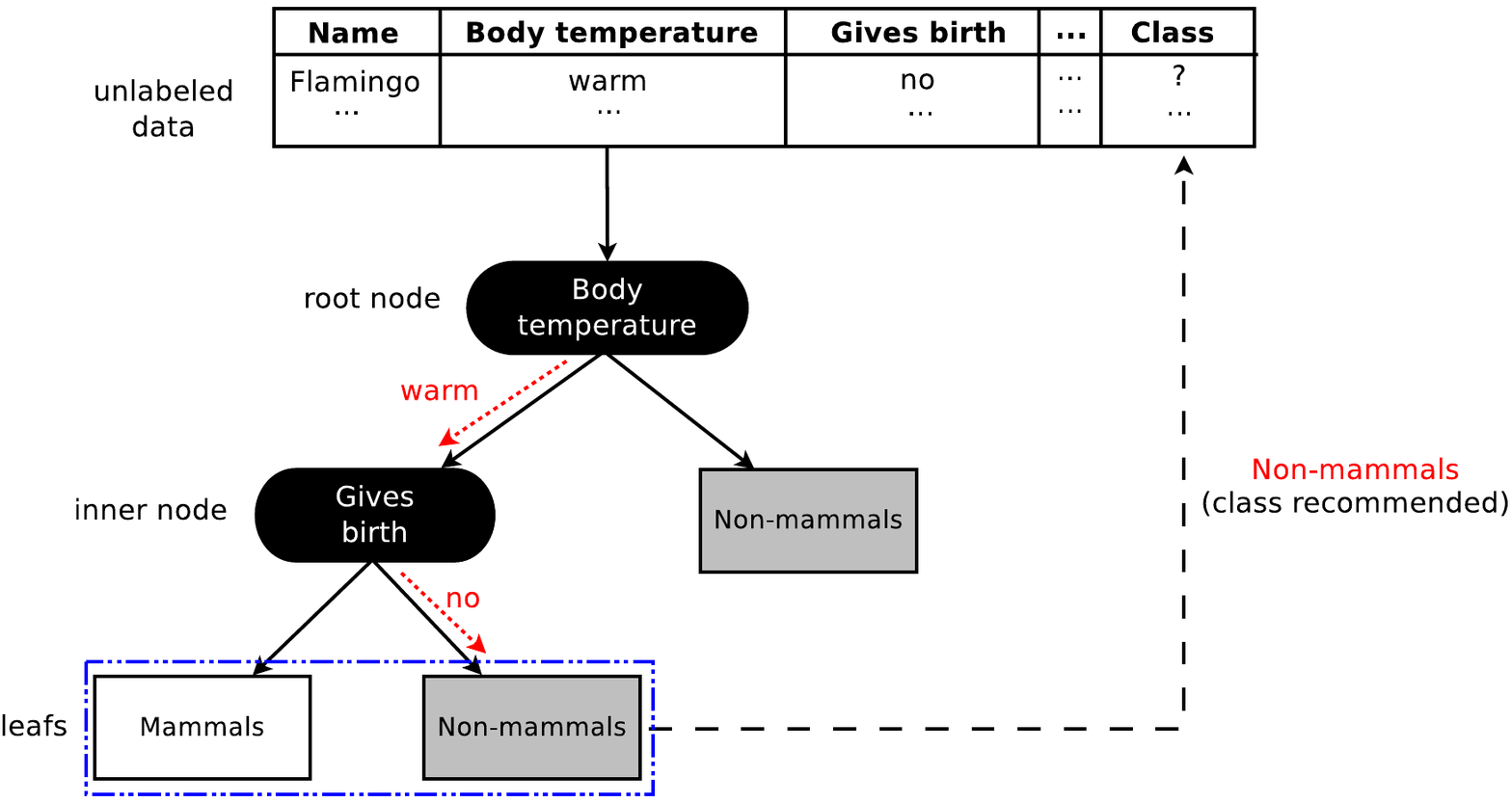}
    \caption{Example of a decision tree classification. When unlabeled data is provided to the tree, conditions are applied starting from the root node and following the appropriate branch until a leaf is reached. The class is recommended based on the leaf pointed out. Adapted from~\citet{Tan:2005}.}
    \label{fig:dt}
\end{figure}

\begin{sloppypar}
As \new{a} consequence of the growing concerns regarding the development of responsible and ethical \acrfull{ai} solutions and the attendance of the requirements of new \acrshort{ai}-related legislation, such as the \acrfull{gdpr}~\citep{GDPR:2016}, model interpretability has \new{become an essential} issue in the \acrshort{ai} research agenda.
Thus, when selecting a \acrfull{ml} algorithm for a new classification task,
good predictive performance coupled with easy model interpretation
favors the \acrfull{dt} induction algorithms~\citep{Maimon:2014}. These algorithms induce a model represented by a set of rules in a tree-like structure (as illustrated in Figure~\ref{fig:dt}). This structure elucidates how the induced model predicts the class of a new instance, more interpretable than many other model representations, such as an \acrfull{ann}~\citep{Haykin:2007} or \acrfullpl{svm}~\citep{Abe:2005}. As a result, \acrshort{dt} induction algorithms are among the most frequently used \acrshort{ml} algorithms for classification tasks ~\citep{Jankowski:2014,Wu:2009}.
\end{sloppypar}


\begin{sloppypar}
\acrshort{dt} algorithms have several other advantages over many \acrshort{ml} algorithms, such as robustness to noise, tolerance against missing information, capability to handle various types of attributes, treatment of irrelevant and redundant attributes,  and low computational cost~\citep{Maimon:2014}. Their importance is attested by the wide range of well-known algorithms proposed in the literature, such as Breiman et al.\textquotesingle s \acrfull{cart}~\citep{Breiman:1984} and Quinlan\textquotesingle s C4.5 algorithm~\citep{Quinlan:1993}, as well as some hybrid-variants of them, like \acrfull{nbtree}~\citep{Kohavi:1996}, \acrfull{lmt}~\citep{Landwehr:2005} and \acrfull{ctree}~\citep{Hothorn:2006}, to name a few.
\end{sloppypar}

\begin{sloppypar}
Similarly to most \acrshort{ml} algorithms, \acrshort{dt} induction algorithms might improve their performance through hyperparameters setup. Due to the high number of possible configurations and their \new{significant} influence on the predictive performance of the induced models, hyperparameter tuning is often warranted~\citep{Bergstra:2011,Massimo:2016,Pilat:2013,Padierna:2017}. Moreover, \new{highly} accurate \acrshort{dt} induction algorithms grounded on pre-prune (e.g., CTree and CHAID) or post-prune (e.g., C4.5, J48 and CART) strategies demand setting up appropriate hyperparameter settings since inaccurate ones lead to under-fitting or over-fitting problems \citep{loh2014fifty}. The tuning task is usually performed to ``black-box'' algorithms, such as \acrshortpl{ann} and \acrshortpl{svm}, but not for \acrshortpl{dt}. There are some prior studies investigating the evolutionary design of new \acrshort{dt} induction algorithms~\citep{Barros:2012,Barros:2015}, but only a few on hyperparameter tuning~\citep{Molina:2012,Reif:2011,Reif:2014}.

Alternatively, the default hyperparameter values suggested in the \new{vastest} most vast of \acrshort{dt} implementations might support effective predictive performance in ordinary classification problems. Furthermore, the tuning execution \new{does not guarantee} improvements over the default values at the cost of extra computational effort.
\end{sloppypar}

Shedding light on the effects of the hyperparameter tuning, in this paper, we investigated the hyperparameter profile of \acrshort{dt} induction algorithms by evaluating their predictive performance and analyzing their convergence during the tuning procedure. We conducted experiments to answer the following research questions:
\begin{enumerate}[i)]
    \item \new{Is \acrshort{hp} tuning of \acrshort{dt} really necessary}?
    \item When performing  \acrshort{hp} tuning, which are the most recommended techniques to perform such a task regarding predictive performance and computational time?
    \item Which  \acrshortpl{hp} most impact \acrshort{dt} predictive performance?
    \item When to tune?
\end{enumerate}


For such, two \acrshort{dt} of the most popular decision tree induction algorithms~\citep{Wu:2009} were chosen as study cases: the J48 algorithm, a \texttt{WEKA}~\citep{Witten:2005} implementation for the Quinlan`s C4.5~\citep{Quinlan:1993}; and the \citet{Breiman:1984}\textquotesingle s \acrshort{cart} algorithm~\cite{Breiman:1984}.
A total of six different hyperparameter tuning techniques (following different learning biases) were benchmarked in the experiments: a simple \acrfull{rs}, three commonly used meta-heuristics - \acrfull{ga}~\citep{Goldberg:1989}, \acrfull{pso}~\citep{Kennedy:1995}, and \acrfull{eda}~\citep{Hauschild:2011}, \acrfull{smbo} ~\citep{Snoek:2012}, and \acrfull{irace}~\citep{Birattari:2010}\footnote{These techniques will be described on the \new{following} sections.}. Experiments were carried out with a large number of heterogeneous datasets, and the experimental results obtained by these optimization techniques are compared with those obtained using the default hyperparameter values recommended for C4.5 and CART. 

In addition, we also assess the relative importance of \acrshort{dt} hyperparameters, measured using a recent functional ANOVA framework~\citep{Hutter:2014}.
In all, the main contributions of this study are:
\begin{enumerate}[i)]
    \item Provide further evidence of the need to perform HP tuning for \acrshort{dt} for some cases;
    \item Large-scale comparison of different hyperparameter tuning techniques for \acrshort{dt} induction algorithms; 
    \item Comprehensive analysis of the hyperparameter profile of \acrshort{dt} induction algorithms, especially the effect of hyperparameters on the predictive performance of the induced models and the relationship between them;
    \item Present simple rules to users \new{to} decide when to tune \acrshort{dt} HPs.
\end{enumerate}
\noindent All the code generated in this study is available to reproduce our analysis and extend it to other classifiers. All experiments are also available on OpenML~\citep{Vanschoren:2014}.

The current study collectively highlights the importance of using \acrshortpl{dt} for interpretability and explainability in \acrshort{ml} models. Literature also reports recent research in this direction. For example, \citet{Blanco-Justica:2019, Blanco-Justica:2020} propose using \acrshortpl{dt} as surrogate models to explain black-box models, achieving a trade-off between comprehensibility and representativeness. \citet{Vieira:2020} focus on understanding the decisions made by \acrshortpl{svm} classifiers using \acrshortpl{dt} as interpretable models. \citet{Ribeiro:2016} argue for model-agnostic interpretability approaches, treating \acrshort{ml} models as black-box functions and providing flexibility in model choice, explanations, and representations. These papers emphasize the need for explainability in critical \acrshort{ai} applications and the benefits of \acrshort{dt} in achieving interpretability and trust in \acrshort{ml} models.

The remainder of this paper is structured as follows: Section~\ref{sec:related} covers related work on hyperparameter tuning of \acrshort{dt} induction algorithms, and Section~\ref{sec:tuning} introduces hyperparameter tuning in more detail. Section~\ref{sec:methods} describes our experimental methodology and the setup of the tuning techniques used, after which Section~\ref{sec:results} analyses the results. 
Section~\ref{sec:validity} validates the results \new{of} this study.
Finally, Section~\ref{sec:conclusions} summarizes our findings and highlights future avenues of research.


\section{Related work} \label{sec:related}

\begin{sloppypar}
\new{Many} ML studies investigate the effect of hyperparameter tuning on the predictive performance of classification algorithms. Most of them deal with the tuning of ``black-box'' algorithms, such as \acrshortpl{svm}~\citep{Gomes:2012} and \acrshortpl{ann}~\citep{Bergstra:2012}; or ensemble algorithms, such as \acrfull{rf}~\citep{Reif:2012,Huang:2016} and Boosting Trees~\citep{Eggensperger:2015,Wang:2015}. They often tune the hyperparameters by using simple techniques, such as \acrfull{ps}~\citep{Eitrich:2006} and \acrfull{rs}~\citep{Bergstra:2012}, but also more sophisticated ones, such as meta-heuristics~\citep{Gascon-Moreno:2011,Gomes:2012,Nakamura:2014, Ridd:2014,Padierna:2017}, \acrshort{smbo}~\citep{Bergstra:2011,Bardenet:2013}, racing algorithms~\citep{Lang:2013,Miranda:2014} and \acrfull{mtl}~\citep{Feurer:2015}. However, when considering \acrshort{dt} induction algorithms, \new{far fewer studies are} available.
\end{sloppypar}
 
Recent work has also used meta-heuristics to design new \acrshort{dt} induction algorithms combining components of existing ones~\citep{Barros:2015,Podgorelec:2015}. 
\new{The existing components restrict the algorithms created}, and since they have to optimize the algorithm and its hyperparameters, they have a much larger search space and computational cost. Since this study focuses on hyperparameter tuning, this section does not cover \acrshort{dt} induction algorithm design.


\subsection{C4.5/J48 hyperparameter tuning}
\label{sec:rel_c45}

Table~\ref{tab:related_J48} summarizes studies performing hyperparameter tuning for the C4.5/J48 \acrshort{dt} induction algorithm. For each study, the table presents which hyperparameters were investigated (following the J48 nomenclature also presented in Table~\ref{tab:hyperparameters})\footnote{The original J48 nomenclature may also be consulted at \url{http://weka.sourceforge.net/doc.dev/weka/classifiers/trees/J48.html.}}, which tuning techniques were explored, and the number and source of datasets used in the experiments. Empty fields in the table mean that the procedures used in that specific study could not be completely identified.

\afterpage{%
    \clearpage
    \thispagestyle{empty}
    \begin{landscape}
        \centering 
            
           \scriptsize
\centering
\captionof{table}{Some properties of the related studies that performed C4.5 (J48) hyperparameter tuning. The hyperparameters abbreviations are explained according to the reference description at the text.}

\label{tab:related_J48}
\setlength{\tabcolsep}{1.1pt}
\begin{tabular}{lcccccccccccccc}
    
    \toprule

    \multicolumn{1}{c}{\multirow{2}{*}{\textbf{Reference}}} & \multirow{2}{*}{Year\:\:}
    & \multicolumn{10}{c}{\textbf{Hyperparameter}} & \textbf{Tuning} & \multicolumn{2}{c}{\textbf{Datasets}} \\ 
    
    & & \textbf{  C  } & \textbf{  M  } & \textbf{  N  } & \textbf{  O  } & \textbf{  R  } & \textbf{  B  } & \textbf{  S  } & \textbf{  A  } & \textbf{  J  } & \textbf{  U  } & \textbf{ Technique } & \multicolumn{1}{c}{\textbf{Number}} & \multicolumn{1}{c}{\textbf{Source}} \\ 
    
    \midrule
    \rule{0pt}{4ex}
    
    \citet{Schauerhuber:2008} & 2008 & $\bullet$ & $\bullet$ & & & & & & & & & GS & 18 & \acrshort{uci} \\
    \rule{0pt}{4ex}
   
    \citet{Sureka:2008} & 2008 & & & & & & & & & & & \acrshort{ga} & - & - \\
    \rule{0pt}{4ex}
        
    \citet{Stiglic:2012} & 2012 & $\bullet$ & $\bullet$ & & & & $\bullet$ & $\bullet$ & & & & \acrshort{vtj48} & 71 & \acrshort{uci} \\
    \rule{0pt}{4ex}
     
    \citet{Lin:2012} & 2012 & $\bullet$ & $\bullet$ & & & & & & & & & SS & 23 & \acrshort{uci} \\ 
    \rule{0pt}{4ex}
    
    \citet{Ma:2012} & 2012 & $\bullet$ & $\bullet$ & & & & & & & & & \acrshort{gp} & 70 & \acrshort{uci} \\ 
    \rule{0pt}{4ex}
    
    \citet{Molina:2012} & 2012 & $\bullet$ & $\bullet$ & & & & & & & & & \acrshort{gs} & 14 & - \\
    \rule{0pt}{4ex}
    
    \citet{Sun:2013} & 2013 & & & & & & $\bullet$ & & & & & \acrshort{pso} & 466 & -\\
    \rule{0pt}{4ex}

    \citet{Reif:2014} & 2014 & $\bullet$ & & & & & & & & & & \acrshort{gs} & 54 &  \acrshort{uci} \\
    \rule{0pt}{4ex}
    
    \citet{Delgado:2014} & 2014 & \multirow{2}{*}{$\bullet$} &  \multirow{2}{*}{$\bullet$} & & & & & & & & & \multirow{2}{*}{-} & \multirow{2}{*}{121} &  \multirow{2}{*}{\acrshort{uci}} \\
   \:\citet{Wainberg:2016} & 2016 \\
      \rule{0pt}{4ex}

    \citet{Kotthoff:2016} & 2016 & $\bullet$ & $\bullet$ & & $\bullet$ & & $\bullet$ & $\bullet$ & $\bullet$ & $\bullet$ & $\bullet$ & \acrshort{smbo} & 21 & - \\ 
    \rule{0pt}{4ex}
    
    \citet{Sabharwal:2016} & 2016 &  $\bullet$ & $\bullet$ & & & & & & & & & DAUP & 6 & -  \\ 
    \rule{0pt}{4ex}
    
   \citet{Tantithamthavorn:2016} & 2016 & $\bullet$ & & & & & & & & & & caret & 18 & - \\
   
    \bottomrule
    
\end{tabular}

            \newpage

  \scriptsize
\centering
\captionof{table}{Summary of previous studies on \acrshort{cart} hyperparameter tuning. The hyperparameter nomenclature adopted is explained according to the reference description in the original text.} 

\label{tab:related_CART}
\setlength{\tabcolsep}{1.1pt}
\begin{tabular}{lccccccccccc}

    \toprule

    \multicolumn{1}{c}{\multirow{3}{*}{\textbf{Reference}}} & & \multicolumn{7}{c}{\textbf{Hyperparameter}} & \textbf{Tuning} & \multicolumn{2}{c}{\textbf{Datasets}} \\ 
    
    &\textbf{Year}\:\: & \textbf{cp } & \textbf{min} & \textbf{min} & \textbf{max} & \textbf{weights} & \textbf{max} & \textbf{max} & \textbf{ Technique } & \multicolumn{1}{c}{\textbf{Number}} &  \multicolumn{1}{c}{\textbf{Source}} \\ 
    
    & & & \textbf{ split } & \textbf{ bucket } & \textbf{ depth } & \textbf{ leaf } & \textbf{ leaf } & \textbf{ feat } & & & \\

    \midrule
    \rule{0pt}{4ex}
    
    \citet{Schauerhuber:2008} & 2008 & $\bullet$ & & & & & & & GS & 18 & \acrshort{uci} \\
    \rule{0pt}{4ex}
   
    \citet{Sun:2013} & 2013 & & $\bullet$ & & & & & & \acrshort{pso} & 466 & - \\
    \rule{0pt}{4ex}
    
    \citet{Delgado:2014} & 2014 &  \multirow{2}{*}{$\bullet$} & & & \multirow{2}{*}{$\bullet$} & & & & \multirow{2}{*}{-} &  \multirow{2}{*}{121} &   \multirow{2}{*}{\acrshort{uci}} \\
   \;\citet{Wainberg:2016} & 2016 \\
    \rule{0pt}{4ex}
    
    \multirow{3}{*}{\citet{Chacon:2015}} & &  & \multirow{3}{*}{$\bullet$} & \multirow{3}{*}{$\bullet$} & \multirow{3}{*}{$\bullet$} & \multirow{3}{*}{$\bullet$} & \multirow{3}{*}{$\bullet$} & \multirow{3}{*}{$\bullet$} & \acrshort{rs} & \\
     &  2015 & & & & & & & & \acrshort{sh} & 36 & \acrshort{uci} \\
     & & & & & & & & & \acrshort{pd} &  \\    
    \rule{0pt}{4ex}

    \citet{Feurer:2015B}& 2015 & & $\bullet$ & & $\bullet$ & $\bullet$ & $\bullet$ & $\bullet$ & \acrshort{smbo} & 140 & \acrshort{openml} \\
    \rule{0pt}{4ex}

   \citet{Levesque:2016} & 2016 & & $\bullet$ & $\bullet$ & $\bullet$ & & $\bullet$ & & \acrshort{smbo} & 18 & \acrshort{uci} \\
    \rule{0pt}{4ex}

   \citet{Tantithamthavorn:2016} & 2016 & $\bullet$ & & & & & & & caret & 18 & various \\
    \rule{0pt}{4ex}

    \textcolor{black}{\citet{Probst:2019}} & 2019 & \textcolor{black}{$\bullet$} & \textcolor{black}{$\bullet$} & \textcolor{black}{$\bullet$} & \textcolor{black}{$\bullet$} & & & & \textcolor{black}{\acrshort{rs}} & \textcolor{black}{38} & \textcolor{black}{\acrshort{openml}} \\
    \rule{0pt}{4ex}

    \textcolor{black}{\citet{Bartz:2021}} & 2021 & \textcolor{black}{$\bullet$} & \textcolor{black}{$\bullet$} & \textcolor{black}{$\bullet$} & \textcolor{black}{$\bullet$} & & & & \textcolor{black}{\acrshort{smbo}} & \textcolor{black}{1} & \textcolor{black}{\acrshort{openml}} \\
    
    \bottomrule
\end{tabular}

    \end{landscape}
    \clearpage
}


\citet{Schauerhuber:2008} presented a benchmark of four different open-source \acrshort{dt} induction algorithm implementations, one being J48. \new{This study assessed} the algorithm's performances on $18$ classification datasets from the \acrshort{uci} repository. The authors tuned two hyperparameters: the pruning confidence (\texttt{C}) and the minimum number of instances per leaf (\texttt{M}).

\citet{Sureka:2008} used a \acrshort{ga} (see Section~\ref{sub:ga}) to recommend an algorithm and its best hyperparameter values for a problem. They used a binary representation to encode a wider hyperparameter space, including Bayes, Rules, Network, and Tree-based algorithms, including J48. However, the authors do not provide more information about which hyperparameters, ranges, datasets, or evaluation procedures were used to assess the hyperparameter settings. Experiments showed that the algorithm can find good solutions but requires massive computational resources to evaluate all possible models.

\citet{Stiglic:2012} presented a study tuning a \acrfull{vtj48}, i.e., J48 with predefined visual boundaries. They developed a new adapted binary search technique to perform the tuning of four J48 hyperparameters: the pruning confidence (\texttt{C}); the minimum number of instances per leaf (\texttt{M}); the use of binary splits (\texttt{B}) and subtree raising (\texttt{S}). Experimental results on $40$ \acrshort{uci}~\citep{Bache:2013} and $30$ bioinformatics datasets demonstrated a significant increase in accuracy in visually tuned \acrshort{dt}s, when compared with defaults. In contrast to classical \acrshort{ml} datasets, \new{bioinformatics datasets had higher gains.}

\citet{Lin:2012} proposed a novel \acrfull{ss}-based algorithm to acquire optimal hyperparameter settings and to select a subset of features that results in better classification performance. Experiments with $23$ \acrshort{uci} datasets demonstrated that the hyperparameter settings for C4.5 algorithm obtained by the new approach, when tuning the `\texttt{C}' and `\texttt{M}' hyperparameters, were better than those obtained by baselines (defaults, simple \acrshort{ga} and a greedy combination of them). When feature selection is considered, \new{most datasets' classification accuracy rates increase.}

\citet{Ma:2012} leveraged the \acrfull{gp} algorithm to optimize hyperparameters for some \acrshort{ml} algorithms (including C4.5 and its hyperparameters `\texttt{C}' and `\texttt{M}') for $70$ \acrshort{uci} classification and regression datasets. \acrshort{gp}s were compared with \acrfull{gs} and \acrshort{rs} methods (see Section~\ref{sub:rs}). \acrshort{gp}s found solutions faster than both baselines with comparably high performances. However, compared specifically to \acrshort{rs}, \acrshort{gp}s seems to be better for more complex problems, while \acrshort{rs} is sufficient for simpler ones.

\citet{Sabharwal:2016} proposed a method to sequentially allocate small data batches to selected \acrshort{ml} classifiers. The method called ``Data Allocation using Upper Bounds'' (DAUP) tries to project an optimistic upper bound of the accuracy obtained by a classifier in the entire dataset, using recent evaluations of this classifier on small data batches. Experiments evaluated the technique on $6$ classification datasets and more than $40$ algorithms with different hyperparameters, including C4.5 and its `\texttt{C}' and `\texttt{M} ' hyperparameters. The proposed method was able to select near-optimal classifiers with a meager computational cost compared to full training of all classifiers.

\citet{Tantithamthavorn:2016}, investigated the performance of prediction models when tuning hyperparameters using ``\texttt{caret}''\footnote{\url{https://cran.r-project.org/web/packages/caret/index.html}}~\citet{caret:2016}, a \acrshort{ml} tool. ML algorithms, including J48 and its `\texttt{C}' hyperparameter, were tuned on $18$ proprietary and public datasets. In a comparison with defaults from \texttt{caret} using the AUC\footnote{Area under the ROC curve} measure, the tuning produced better results.

\begin{sloppypar}
\citet{Wainberg:2016} reproduced the benchmark experiments described in~\citet{Delgado:2014}. 
They evaluated $179$ classifiers from $17$ different learning groups on $121$ datasets from \acrshort{uci}. The hyperparameters of the J48 algorithm were manually tuned.
\end{sloppypar}


Other studies used hyperparameter tuning methods to generate \acrfull{mtl} systems~\citep{Molina:2012,Sun:2013,Reif:2014,Kotthoff:2016}. These studies search the hyperparameter spaces to describe the behavior of \acrshort{ml} algorithms in a set of problems and later recommend hyperparameter values for new problems. For example, Molina et. al.~\cite{Molina:2012} tuned two hyperparameters of the J48 algorithm (`\texttt{C}' and `\texttt{M}') in a case study with $14$ educational datasets, using  \acrshort{gs}. They also used a set of meta-features to recommend the most promising set of algorithms and \acrshortpl{hp} for each problem. The proposed approach, however, did not improve the performance of the \acrshort{dt}s with defaults.

\citet{Sun:2013} also used hyperparameter tuning in the context of \acrshort{mtl}. The authors proposed a new meta-learner for algorithm recommendation and a feature generator to construct the datasets used in experiments. They searched $20$ \acrshort{ml} algorithm hyperparameter spaces, one of them the C4.5 and its `\texttt{B}' hyperparameter. The \acrshort{pso} technique (see Section~\ref{sub:pso}) was used to generate a meta-database for a recommendation experiment. Similarly, \citet{Reif:2014} implemented an open-source \acrshort{mtl} system to predict accuracies of target classifiers, one of them the C5.0 algorithm (a version of the C4.5), with its pruning confidence (\texttt{C}) tuned by \acrshort{gs}.

A special case of hyperparameter tuning is the \acrfull{cash} tool, introduced by \citet{Thornton:2013} as the Auto-WEKA\footnote{\url{http://www.cs.ubc.ca/labs/beta/Projects/autoweka/}} framework, and updated recently in \citet{Kotthoff:2016}. Auto-WEKA applies \acrshort{smbo} (see Section~\ref{sub:smbo}) to select an algorithm and its hyperparameters to new problems based on a comprehensive set of \acrshort{ml} algorithms (including J48). In addition to the previously mentioned hyperparameters (\texttt{C}, \texttt{M}, \texttt{B} and \texttt{S}), Auto-WEKA also searches for the following HP values: whether to collapse the tree (\texttt{O}), use of Laplace smoothing (\texttt{A}), use of MDL correction for the info gain criterion (\texttt{J}) and generation of unpruned trees (\texttt{U}).


\subsection{CART hyperparameter tuning}
\label{sec:rel_cart}

Table~\ref{tab:related_CART} summarizes previous studies on hyperparameter tuning for the \acrshort{cart} algorithm. \new{The table presents which hyperparameters, tuning techniques, and the number and source of datasets explored in the experiments for each study.}

In \citet{Schauerhuber:2008}, the authors added \acrshort{cart}/rpart to their benchmark analysis. They manually tuned only the complexity parameter `\texttt{cp}'. Sun et. al.~\cite{Sun:2013} investigated the tuning of \acrshort{cart} hyperparameters, in particular its \texttt{minsplit} hyperparameter, over $466$ datasets (some of which are artificially generated) using \acrshort{pso}. This hyperparameter controls the minimum number of instances necessary for a split to be attempted. 
The hyperparameter settings assessed during the search were used to feed a meta-learning system. In~\citet{Tantithamthavorn:2016}, the authors did a similar study but focused on the complexity parameter `\texttt{cp}'.

In~\citet{Chacon:2015}, the authors presented a hierarchical model selection framework that automatically selects the best \acrshort{ml} algorithm for a particular dataset, optimizing its hyperparameter values. 
Algorithms and hyperparameters are organized in a hierarchy, and an iterative process makes the recommendation. The optimization technique used for tuning is considered a component of the framework, and three choices are available: \acrshort{rs}, \acrfull{sh}, and \acrfull{pd} optimization methods.
The technique encapsulates a long list of algorithms, including \acrshort{cart} and some of its hyperparameters: `\texttt{minsplit}'; the minimum number of instances in a leaf (`\texttt{minbucket}'); the maximum depth of any node of the final tree (`\texttt{maxdepth}'); weighted values to leaf nodes (`\texttt{weights\_leaf}'); the maximum number of leaves (`\texttt{maxleafs}') and the maximum number of features from dataset used in trees (`\texttt{maxfeatures}').

In~\citet{Feurer:2015B}, the authors used the \acrshort{smbo} approach to select and tune algorithm from the ``\texttt{scikit learn}''\footnote{\url{http://scikit-learn.org/}} framework, hence Auto-skLearn\footnote{\url{https://github.com/automl/auto-sklearn}}. The only \acrshort{dt} induction algorithm covered here is \acrshort{cart}. \acrshort{cart} with some hyperparameters manually selected was also experimentally investigated in~\citet{Delgado:2014} and \citet{Wainberg:2016}.

\citet{Levesque:2016} investigated the use of hyperparameter tuning and ensemble learning for tuning \acrshort{cart} hyperparameters when models induced by \acrshort{cart} were part of an ensemble, using \acrshort{smbo}. Four hyperparameters were tuned in the process: `\texttt{minsplit}', `\texttt{minbucket}', `\texttt{maxdepth}' and the `\texttt{maxleaf}'. The tuning resulted in a significant improvement in generalization accuracy when compared with the Single Best Model Ensemble and Greedy Ensemble Construction techniques. 

\citet{Probst:2019} formalize the problem of tuning from a statistical point of view. They have conducted experiments with $38$ datasets from \acrshort{openml} and six common \acrshort{ml} algorithms, one of them is \acrshort{cart}, and tuning all of them using a \acrshort{rs} technique. Results reported enable users to decide whether conducting a possibly time-consuming tuning strategy is worthwhile. \citet{Bartz:2021} have performed a similar experiment, but considering a single dataset: CID from \acrshort{openml}. In this study, the authors' focus was on providing useful scripts in R that could be used by experts and non-experts when performing tuning of \acrshort{ml} algorithms.


\subsection{Literature Overview}
\label{sec:gap}

The literature review indicates that hyperparameter tuning for \acrshort{dt} induction algorithm could be more deeply explored. We found eleven studies investigating tuning for the J48 algorithm and nine for \acrshort{cart}. These studies neither investigated the tuning task itself nor adopted a consistent procedure to assess candidate hyperparameter settings while searching the hyperparameter space:
\begin{itemize}
    \item some studies used hyperparameter sweeps;
    \item some other studies used simple \acrshort{cv} resamplings;
    \item a few studies used nested-\acrshort{cv} procedures, but only used an inner holdout, and they did not repeat their experiments with different seeds\footnote{Since the stochastic nature of the often used tuning algorithms, experimenting with different seeds (for random generator) is desirable.}; and
    \item some studies did not even describe the experimental methodology used. 
\end{itemize}

Regarding the search space, most studies concerning C4.5/J48 and \acrshort{cart} hyperparameter tuning investigated only a small subset of the hyperparameter search spaces (as shown in Tables~\ref{tab:related_J48} and \ref{tab:related_CART}). Furthermore, most of the studies did the tuning manually, used simple hyperparameter tuning techniques, or searched the hyper-spaces to generate meta-information for \acrfull{mtl} and \acrshort{cash} systems. 

This paper overcomes these limitations by investigating several techniques for \acrshort{dt} hyperparameter tuning using a reproducible and consistent experimental methodology. It also analyzes the importance and relationships between many hyperparameters of the investigated algorithms (C4.5 and \acrshort{cart}).
These results increase awareness of the importance of tuning \acrshort{dt} and provide guidance on performing this task considering two critical criteria: improve predictive performance while minimizing computation cost.


\section{Hyperparameter tuning}
\label{sec:tuning}

Many applications of \acrshort{ml} algorithms to classification tasks use hyperparameter default values suggested by \acrshort{ml} tools, even though several studies have shown that their predictive performance \new{mainly} depends on using the right hyperparameter values~\citep{Feurer:2015B, Thornton:2013}. 
In early works, these values were tuned according to previous experiences or by trial and error. 
Depending on the training time available, finding a good set of values manually may be subjective and time-consuming. In order to overcome this problem, optimization techniques are often employed to automatically look for a suitable set of hyperparameter settings~\citep{Bergstra:2011}.

The hyperparameter tuning process is usually treated as a black-box optimization problem whose objective function is associated with the predictive performance of the model induced by a \acrshort{ml} algorithm, formally defined as follows.


Let $\mathcal{H}=\mathcal{H}_1\times\mathcal{H}_2\times\dots\times\mathcal{H}_k$ be the hyperparameter space for an algorithm $a\in\mathcal{A}$, where $\mathcal{A}$ is the set of \acrshort{ml} algorithms. Each $\mathcal{H}_i$ represents a set of possible values for the $i^{th}$ hyperparameter of $a$ ($i\in\{1, \dots, k\}$) and can be usually defined by a set of constraints.
Additionally, let $\mathcal{D}$ be a set of datasets where $\mathbf{D}\in\mathcal{D}$ is a dataset from $\mathcal{D}$.
The function $f:\mathcal{A}\times\mathcal{D}\times\mathcal{H}\rightarrow\mathbb{R}$ measures the predictive performance of the model induced by the algorithm $a\in\mathcal{A}$ on the dataset $\mathbf{D}\in\mathcal{D}$ given a hyperparameter configuration $\mathbf{h}=(h_1,h_2,\dots, h_k)\in\mathcal{H}$. Without loss of generality, higher values of $f$ mean higher predictive performance.

Given an algorithm $a\in\mathcal{A}$, its hyperparameter space $\mathcal{H}$ and a dataset $\mathbf{D}\in\mathcal{D}$, 
the goal of hyperparameter tuning is to find $\mathbf{h}^\star=(h_1^\star,h_2^\star,\dots, h_k^\star)$ such that

\begin{equation}
\label{eq:setup}
      \centering
  \mathbf{h}^\star = \underset{\mathbf{h}\in\mathcal{H}}{arg\:max}\:f(a,\mathbf{D},\mathbf{h})
\end{equation}


The optimization of the hyperparameter values can be based on any performance measure $f$, which can even be defined by multi-objective criteria. Further aspects can make the tuning more difficult, \new{such as}:
\begin{itemize}
    \item hyperparameter configurations that lead to a model with high predictive performance for a given dataset may not lead to high predictive performance for other datasets;
    \item hyperparameters often depend on each other (as in the case of \acrshortpl{svm}~\citep{BenHur:2010}). Hence, independent tune of hyperparameters may not lead to good settings; 
    \item the evaluation of a specific hyperparameter configuration, not to mention many configurations, can be subjective and very time-consuming.
\end{itemize}

In the last decades, population-based optimization techniques have been successfully used for hyperparameter tuning of classification algorithms~\citep{Bardenet:2013}. When applied to tuning, these techniques (iteratively) build a \textit{population} $\mathcal{P}\subset\mathcal{H}$ of hyperparameter settings for which $f(a,\mathbf{D},\mathbf{h})$ are being computed for each $\mathbf{h}\in\mathcal{P}$. By doing so, they can simultaneously explore different regions of a search space. There are various population-based hyperparameter tuning strategies, which differ in how they update $\mathcal{P}$ at each iteration. 
Some of them are briefly described next.


\subsection{Random Search}
\label{sub:rs}

\acrfull{rs} is a simple technique that performs random trials in a search space. Its use can reduce the computational cost when a large number of possible settings are being investigated~\citep{Andradottir:2015}. Usually, \acrshort{rs} performs its search iteratively in a predefined number of iterations. Moreover, $\mathcal{P}_i$ is extended (updated) by a randomly generated hyperparameter setting $\mathbf{h}\in\mathcal{H}$ in each ($i$th) iteration of the hyperparameter tuning process. \acrshort{rs} has obtained efficient results in optimization for \acrfull{dl} algorithms~\citep{Bergstra:2012, Bardenet:2013}.


\subsection{Sequential Model Based Optimization}
\label{sub:smbo}

\acrfull{smbo}~\citep{Snoek:2012} is a sequential method that starts with a small initial population $\mathcal{P}_0\neq\emptyset$ which, at each new iteration $i>0$, is extended by a new hyperparameter configuration $\mathbf{h}'$, such that the expected value of $f(a,\mathbf{D},\mathbf{h}')$ is maximal according to an induced meta-model $\hat{f}$ approximating $f$ on the current population $\mathcal{P}_{i-1}$. In~\citet{Bergstra:2011, Snoek:2012, Bergstra:2013B}, \acrshort{smbo} performed better than \acrshort{gs} and \acrshort{rs} and matched or outperformed state-of-the-art techniques in several hyperparameter optimization tasks.


\subsection{Genetic Algorithm} 
\label{sub:ga}

\begin{sloppypar}
Bio-inspired techniques, such as a \acrfull{ga}, based on natural processes, have also been largely used for hyperparameter tuning~\citep{Gomes:2012}. In these techniques, the initial population $\mathcal{P}_0=\{\mathbf{h}_1,\mathbf{h}_2,\dots,\mathbf{h}_{n_0}\}$, generated randomly or according to background knowledge, is changed in each iteration according to operators based on natural selection and evolution.
\end{sloppypar}


\subsection{Particle Swarm Optimization} 
\label{sub:pso}

\acrfull{pso} is a bio-inspired technique relying on the swarming and flocking behaviors of animals~\citep{Simon:2013}. In case of \acrshort{pso}, each particle $\mathbf{h}\in\mathcal{P}_0$ is associated with its position $\mathbf{h}=(h_1,\dots,h_k)\in\mathcal{H}$ in the search space $\mathcal{H}$, a velocity $\mathbf{v}_h\in\mathbb{R}^k$ and also its so far best found position $\mathbf{b}_h\in\mathcal{H}$.
During iterations, the movements of each particle \new{are} changed according to its so far best-found position as well as the so far best-found position $\mathbf{w}\in\mathcal{H}$ of the entire swarm (recorded through the computation).


\subsection{Estimation of Distribution Algorithm} 
\label{sub:eda}

\acrfull{eda}~\citep{Hauschild:2011} lies on the boundary of \acrshort{ga} and \acrshort{smbo} by combining the advantages of both approaches such that the search is guided by iteratively updating an explicit probabilistic model of promising candidate solutions. In other words, the implicit crossover and mutation operators used in \acrshort{ga} are replaced by an explicit probabilistic model $M$. 


\subsection{Iterated F-Race} 
\label{sub:irace}

The \acrfull{irace}~\citep{Birattari:2010} technique was designed for algorithm configuration and optimization problems~\citep{Lang:2013, Miranda:2014} based on '\textit{racing}'. One race starts with an initial population $\mathcal{P}_{0}$, iteratively selects the most promising candidates considering the hyperparameter distributions and compares them by statistical tests. Configurations statistically worse than at least one of the other configuration candidates are discarded from the racing. Based on the surviving candidates, the distributions are updated. This process is repeated until a stopping criterion is reached.


\subsection{Other recent techniques}

Recently, new optimization/tuning techniques have also been proposed. One that stands out is Hyperband~\citep{hyperband:2018}.
Hyperband is an early-stopping method that adaptively allocates some predefined resource (iterations, data samples, or features) to randomly sampled configurations. Then, models are trained with each configuration, and the technique stops training configurations that perform poorly while allocating additional resources to promising configurations. The key concept here is the successive halving: half of the configurations are thrown out at each iteration based on performance. The top-best are kept and trained with a new budget. The process is repeated until one configuration remains.

In~\citet{Falkner:2018}, the authors proposed a new technique called ~\acrfull{bohb}, combining Bayesian Optimization and Hyperband. Unlike Hyperband, which proposes hyperparameter configurations randomly, \acrshort{bohb} uses a model-based approach equivalent to maximizing expected improvement. The authors also showed empirically that \acrshort{bohb} performs similarly to Hyperband in low-budget executions but outperforms both methods (\acrshort{smbo} and Hyperband) when enough budget are available\footnote{For a complete survey on hyperparameter tuning techniques and perspectives, please, consult~\citet{Bischl:2023}.}.



\section{Experimental methodology} 
\label{sec:methods}

\begin{figure*}[h!]
    \centering
    \includegraphics[width = \textwidth]{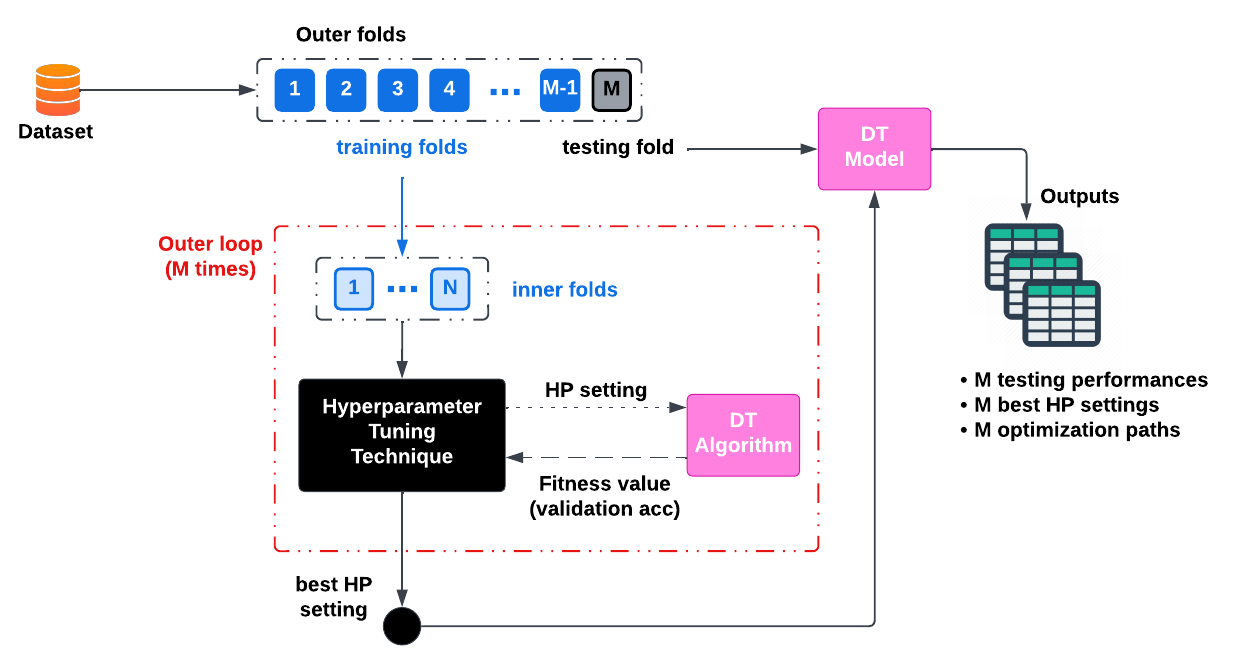}
    \caption{Experimental methodology used to adjust \acrshort{dt} hyperparameters. The tuning is conducted via nested cross-validation: $3$-fold \acrshort{cv} for computing fitness values and $10$-fold \acrshort{cv} for assessing performances. The outputs are the hyperparameter settings, the predicted performances, and the optimization paths of each technique.}
    \label{fig:exp_methods}
\end{figure*}

The experimental methodology employed to analyze the hyperparameter tuning of \acrshort{dt} is illustrated by Figure~\ref{fig:exp_methods}. It follows the nested \acrfull{cv}~\citep{Cawley:2010, Krstajic:2014} resampling procedure. 
For each dataset, data are split into $M$ outer-folds. For each iteration, the tuning techniques use $M-1$ folds to find hyperparameter settings that aim to improve the models' predictive performance, while the remaining fold is used to assess the `optimal' solution found. Internally, the tuning techniques merge the $M-1$ folds and split them to $N$ inner-folds, used in a simple \acrshort{cv} procedure to train the models and assess their predictive performance (fitness) for each hyperparameter setting. At the end of the process, a set of $M$ optimization paths, $M$ settings, and their predictive performances are returned. 
During the experiments, all the tuning techniques were run on the same data partitions, with the same seeds and data to allow their comparison.

In~\citet{Krstajic:2014}, the authors defined $M = N = 10$. However, they argued that there is no rule on choosing the number of folds in the outer and inner \acrshort{cv} loops. 
Here, we have also used $M = 10$, but due to time constraints and the size of datasets used in experiments, $N = 3$ was adopted. The following subsections detail the sub-components used in the tuning task.


\subsection{Hyperparameter spaces} 
\label{subsec:space}

The experiments were performed considering the hyperparameter tuning of two \acrshort{dt} induction algorithms: the `J48' algorithm, a \texttt{WEKA}\footnote{\url{http://www.cs.waikato.ac.nz/ml/weka/}}~\citep{Witten:2005} implementation of the C4.5 algorithm; and the \texttt{rpart} implementation of the
\acrshort{cart}~\citep{Breiman:1984} algorithm.
These algorithms were selected due to their wide acceptance and use in many \acrshort{ml} applications~\citep{Barros:2012,Maimon:2014,Jankowski:2014}. Furthermore, both algorithms are among the most used in \acrshort{ml}, especially by non-expert users~\citep{Wu:2009}.
The correspondent hyperparameter spaces investigated are described in Table~\ref{tab:hyperparameters}.

\afterpage{%
    \clearpage
    \begin{landscape}
        \centering 
    \footnotesize
    \centering
    \captionof{table}{Decision Tree hyperparameter spaces explored in the experiments. The J48 nomenclature is based on the \texttt{RWeka} package, and the \acrshort{cart} terms is based on the \texttt{rpart} package.}
    \label{tab:hyperparameters}
        \begin{tabular}{clllcccc}
            
            \toprule
            
            \textbf{ Algorithm } & \textbf{ Symbol } & \textbf{Hyperparameter} & \textbf{Range} & \textbf{Type} & \textbf{Default} & \textbf{Conditions} \\
            
            \midrule
            \rule{0pt}{1ex}
          
            \multirow{15}{*}{J48} & C & pruning confidence & $(0.001, 0.5) $ & real & 0.25  & R = False \\
            \rule{0pt}{4ex}
          
            & M & minimum number of instances in a leaf & $[1,50]$ & integer & 2 & - \\
            \rule{0pt}{4ex}
            
            & \multirow{2}{*}{N} & number of folds for reduced                       & \multirow{ 2}{*}{$[2,10]$} & \multirow{ 2}{*}{integer} & \multirow{ 2}{*}{3} & \multirow{ 2}{*}{R = True}\\
            & & error pruning & & & & \\
            \rule{0pt}{4ex}
           
            & O & do not collapse the tree & \{False,\:True\}    & logical   & False & - \\
            \rule{0pt}{4ex}
          
            & R & use reduced error pruning & \{False,\:True\}    & logical   & False & - \\
          \rule{0pt}{4ex}
            
            & B & use binary splits only & \{False,\:True\}    & logical   & False & - \\
            \rule{0pt}{4ex}
            
            & S & do not perform subtree raising & \{False,\:True\}    & logical   & False & - \\
            \rule{0pt}{4ex}
           
            & \multirow{ 2}{*}{A} & Laplace smoothing for predicted                & \multirow{ 2}{*}{\{False,\:True\}}    & \multirow{ 2}{*}{logical}   & \multirow{ 2}{*}{False} & - \\
            & &  probabilities & & & & \\
          \rule{0pt}{4ex}
            
            & \multirow{ 2}{*}{J} & do not use MDL correction for & \multirow{ 2}{*}{\{False,\:True\}}    & \multirow{ 2}{*}{logical}   & \multirow{ 2}{*}{False} & - \\
            & & info gain on numeric attributes & & & & \\
            
            \midrule
          
            \rule{0pt}{1ex}
            
            \multirow{12}{*}{CART} & cp & complexity parameter & $(0.0001, 0.1)$ & real & $0.01$ & - \\
            \rule{0pt}{4ex}

            &  \multirow{2}{*}{minsplit} & minimum number of instances in a &  \multirow{2}{*}{$[1,50]$} & \multirow{2}{*}{integer} & \multirow{2}{*}{$20$} &  \multirow{2}{*}{-} \\
            & & node for a split to be attempted \\
             \rule{0pt}{4ex}
          
            & minbucket & minimum number of instances in a leaf & $[1,50]$ & integer & $7$ & - \\
          \rule{0pt}{4ex}
            
            & \multirow{2}{*}{maxdepth} & maximum depth of any node of & \multirow{2}{*}{$[1,30]$} & \multirow{2}{*}{integer} & \multirow{2}{*}{$30$} & \multirow{2}{*}{-} \\
            & & the final tree \\
          \rule{0pt}{4ex}
            
            & \multirow{2}{*}{usesurrogate} & how to use surrogates in the splitting & \multirow{2}{*}{$\{0,1,2\}$} & \multirow{2}{*}{factor} & \multirow{2}{*}{$2$} & \multirow{2}{*}{-} \\
            & & process \\
          \rule{0pt}{4ex}
            
            & \multirow{2}{*}{surrogatestyle} & controls the selection of the best & \multirow{2}{*}{$\{0,1\}$} & \multirow{2}{*}{factor} & \multirow{2}{*}{$0$} & \multirow{2}{*}{-} \\
            & & surrogate \\
            \bottomrule
            
        \end{tabular}
    \end{landscape}
    \clearpage
}


Originally, J48 has ten tunable hyperparameters\footnote{\url{http://weka.sourceforge.net/doc.dev/weka/classifiers/trees/J48.html}}: all presented at Table~\ref{tab:hyperparameters} plus the hyperparameter $U$, which enables the induction of unpruned trees. Since pruned trees look for the most interpretable models without loss of predictive performance, this hyperparameter was removed from the experiments, and just pruned trees were considered. Regarding \acrshort{cart}, all the tunable hyperparameters in \texttt{rpart} were selected.

For each hyperparameter, Table~\ref{tab:hyperparameters} shows the allowed range of values, default values provided by the correspondent R packages, and its constraints for setting new values. The range of values for the hyperparameter $M$ is the same used in~\citet{Reif:2011}. The pruning confidence ($C$) hyperparameter range was adapted from \citet{Reif:2014} because the algorithm internally controls the parameter values, not allowing some values near zero or $C \ge 0.5$.


\subsection{Datasets} 
\label{subsec:datasets}

\begin{sloppypar}
The experiments were carried out using \textbf{$94$} public datasets from the \acrfull{openml}~\citep{Vanschoren:2014} website\footnote{\url{http://www.openml.org/}}, a free scientific platform for standardization of \acrshort{ml} experiments, collaboration and sharing empirical results\footnote{Initially, there were 100 datasets, but 6 of them spent too much time to finish their tuning jobs. They consumed over 1000 hours when we proceeded with their interruption.}. 
Binary and multiclass classification datasets were selected, varying the number of attributes (D) ($3 \leq D \leq 1300$) and examples (N) ($100 \leq N \leq 45000$). In all the selected datasets, each class (C) has at least $10$ examples to allow the use of the stratified methodology. All datasets, with their main characteristics, are presented in Tables~\ref{app:data1}~to~\ref{app:data5} at~\autoref{app:datasetsComplete}.
\end{sloppypar}


\subsection{Hyperparameter tuning techniques} 
\label{subsec:techniques}
 
\begin{sloppypar}
Six hyperparameter tuning techniques were investigated:
\begin{itemize}
    \item three different meta-heuristics: a \acrfull{ga}~\citep{Goldberg:1989}, \acrfull{pso}~\citep{Kennedy:1995} and an \acrfull{eda}~\citep{Hauschild:2011}. These techniques are often used for hyperparameter tuning of \acrshort{ml} classification algorithms in general~\citep{Gascon-Moreno:2011,Yang:2013};
    
    \item a simple \acrfull{rs} technique: suggested by \citet{Bergstra:2012} as a good alternative for hyperparameter tuning replacing \acrfull{gs} technique;
    
    \item \acrfull{irace}~\citep{Birattari:2010}: a \textit{racing} technique designed for algorithm configuration problems; and
    
    \item a \acrfull{smbo}~\citep{Snoek:2012} technique: a state-of-the-art technique for optimization that employs statistical and/or \acrshort{ml} techniques to predict distributions over labels and allows a more direct and faster optimization.

\end{itemize}
\end{sloppypar}

 
\begin{table*}[htb!]
\footnotesize
\centering
\caption{Setup of the hyperparameter tuning experiments.}
\begin{tabular}
    {@{\extracolsep{\fill}}lll}

\toprule
\textbf{Element} & \textbf{Method} & \textbf{R package} \\
\midrule

\multirow{6}{*}{HP-tuning techniques} & Random Search & \texttt{mlr}\\
& Genetic Algorithm   & \texttt{GA}  \\
& Particle Swarm Optimization  & \texttt{PSO} \\
& Estimation of Distribution Algorithm  & \texttt{copulaedas} \\
& Sequential Model Based Optimization & \texttt{mlrMBO} \\
& Iterated F-race & \texttt{irace} \\

\midrule

\multirow{2}{*}{Baseline} & \multirow{2}{*}{Default values (DF)} & \texttt{RWeka} \\
& & \texttt{rpart} \\

\midrule

\multirow{2}{*}{Decision Trees} & J48 algorithm & \texttt{RWeka} \\
 & CART algorithm & \texttt{rpart} \\
 
\midrule

\multirow{2}{*}{Resampling} & Outer: 10-fold \acrshort{cv} & \multirow{2}{*}{\texttt{mlr}} \\
 & Inner: 3-fold \acrshort{cv}  & \\

\midrule

Optimized measure & $\{$\acrfull{bac}$\}$ & \multirow{3}{*}{\texttt{mlr}} \\

\multirow{2}{*}{Evaluation measure} & $\{$\acrfull{bac}, &\\
& Optimization paths $\}$ & \\

\midrule

Budget  & 900 evaluations & - \\

\midrule

\multirow{2}{*}{Repetitions} & 30 times with different seeds & \multirow{2}{*}{-} \\
& seeds = $\{1, \ldots, 30\}$ & \\

\midrule

\multirow{2}{*}{Statistical Evaluation} & Wilcoxon & \multirow{2}{*}{stats} \\
& $\alpha = 0.05$ \\

\bottomrule
 
\end{tabular}
\label{tab:hp_setup}
\end{table*}


\begin{table*}[htb!]
\footnotesize
\centering
\caption{Tuning techniques hyperparameters. Excepting the budget-dependent hyperparameters all of them are the defaults provided by each R package implementation.}
\begin{tabular}
    {@{\extracolsep{\fill}}cll}

\toprule
\textbf{Tuning} & \multicolumn{1}{l}{\multirow{2}{*}{\textbf{Hyperparameters}}} & \multirow{2}{*}{\textbf{Value}} \\
\multicolumn{1}{l}{\textbf{Technique}} \\
\midrule

\multirow{1}{*}{RS} & stopping criteria & budget size \\
\midrule

\multirow{4}{*}{PSO} & number of particles & 10 \\
    & maximum number of iterations & 90 \\
    & stopping criteria & budget size \\
    & algorithm implementation & SPSO2007 \citep{Clerc:2012} \\
\midrule

\multirow{6}{*}{EDA} & number of individuals & 10 \\
    & maximum number of iterations & 90 \\
    & stopping criteria & budget size \\
    & EDA implementation & GCEDA \\
    & copula function & normal \\
    & margin function & truncnorm \\
\midrule

\multirow{9}{*}{GA} & number of individuals & 10 \\
  & maximum number of iterations & 90 \\
  & stopping criteria & budget size \\
  & selection operator & proportional selection with linear scaling  \\
  & crossover operator & local arithmetic crossover \\
  & crossover probability & 0.8 \\
  & mutation operator & random mutation \\
  & mutation probability & 0.05 \\
  & elitism rate & 0.05 \\
\midrule

\multirow{5}{*}{SMBO} & points in the initial design & 10 \\
  & initial design method & Random LHS \\
  & surrogate model & Random Forest \\
  & stopping criteria & budget size \\
  & infill criteria & expected improvement \\
\midrule

\multirow{2}{*}{Irace} & number of instances for resampling & 100 \\
& stopping criteria & budget size \\

\bottomrule

\end{tabular}
\label{tab:tech_setup}

\end{table*}


\begin{sloppypar}
Table~\ref{tab:hp_setup} summarizes the choices to accomplish the general hyperparameter tuning techniques. Experiments were coded and executed with the R language.
Most of the experiments were implemented using the \texttt{mlr} package\footnote{\url{https://github.com/mlr-org/mlr}}~\citep{mlr:2016} (measures, resampling strategies, tuning main processes and \acrshort{rs} technique). The \acrshort{ga}, \acrshort{pso} and \acrshort{eda} meta-heuristics were implemented using the \texttt{GA}\footnote{\url{https://github.com/luca-scr/GA}}\citep{Scrucca:2013}, \texttt{pso}\footnote{\url{https://cran.r-project.org/web/packages/pso/index.html}}\citep{Bendtsen:2012}, and \texttt{copulaedas}\footnote{\url{https://github.com/yasserglez/copulaedas}}\citep{Gonzalez-Fernandez:2014} packages, respectively. 
The J48 and \acrshort{cart} algorithms were implemented using the \texttt{RWeka}\footnote{\url{https://cran.r-project.org/web/packages/RWeka/index.html}}\citep{Hornik:2009}, and \texttt{rpart}\footnote{\url{https://cran.r-project.org/web/packages/rpart/index.html}}\citep{rpart:2014} packages, respectively, wrapped into the \texttt{mlr} package. 
The \acrshort{smbo} technique was implemented using the \texttt{mlrMBO}\footnote{\url{https://github.com/mlr-org/mlrMBO}} package, with its \acrshort{rf} surrogate models implemented by the \texttt{randomForest}\footnote{\url{https://cran.r-project.org/web/packages/randomForest/index.html}} package~\citep{rf:2002}. 
The Irace technique was implemented using the \texttt{irace}\footnote{\url{http://iridia.ulb.ac.be/irace/}}~\citep{irace:2016} package.
\end{sloppypar}


Since the experiments handle \new{many} datasets with different characteristics, many datasets may have unbalanced classes. Thus, the same predictive performance measure used during optimization as the fitness value, \acrfull{bac}~\citep{Brodersen:2010}, is used for model evaluation. 

When tuning occurs in real scenarios, time is a crucial aspect to be considered. Sometimes, the tuning process may take many hours to find good settings for a single dataset~\citep{Reif:2012,Ridd:2014}. 
Thus, this study investigates whether it is possible to find the same good settings faster by using a reduced number of evaluations (budget). 
Based on previous results and analyses~\citep{Mantovani:2016}, a budget size of $900$ evaluations was adopted in the experiments\footnote{The budget size choice is discussed with more details in Section~\ref{sec:validity}.}. 

Since all techniques are stochastic, each one was executed $30$ times for each dataset using different seed values. It gives a total of $270.000 = 30$ (repetitions) $\times\:10$ (outer-folds) $\times \:900$ (budget) HP-settings generated during the search process for one single dataset. So, we evaluated a total of 25.380.000 hyperparameter configurations, considering all the $94$ datasets.
Besides, the default hyperparameter values provided by the `\texttt{RWeka}' and `\texttt{rpart}' packages were used as baseline for the experimental comparisons.

As this paper evaluates different tuning techniques, to avoid the influence of their hyperparameter values on their performances and the recursive problem of tuning the tuning techniques, the authors decided to use their default values.
Each tuning technique has a different set of hyperparameters, which are specific and different considering each technique's paradigm. In the \acrshort{smbo}, \acrshort{irace} and \acrshort{pso} cases, the use of the defaults has been shown robust enough to save time and resources\citep{Bigiarini:2013,irace:2016}. For \acrshort{eda} and \acrshort{ga} (and evolutionary methods in general), there are no \textit{standard} values for their parameters~\citep{Mills:2015}. So, to keep fair comparisons, the default parameter values provided by the correspondent R packages were used. These values may be seen in Table~\ref{tab:tech_setup}.

The tuning techniques have an initial population with $10$ random hyperparameter settings and the same stopping criteria: the budget size~\footnote{The population size = $10$ might be small initially, but it proves to be enough to provide good and accurate results as empirically evaluated in~\citet{Mantovani:2016}.}. The \acrshort{ga}, \acrshort{pso} and \acrshort{eda} techniques use a ``real-value'' codification for the individuals/particles. Thus, they were adapted to handle discrete and Boolean hyperparameters. 
All of them were executed sequentially in the same cluster environment. Every single job generated was executed in a dedicated core with no concurrency and scheduled by the cluster system.

 
\subsection{Repositories for the coding used in this study}
\label{subsec:repo}

Details of the hyperparameter tuning experiments are publicly available in an \acrshort{openml} Study (id $50$). All datasets, classification tasks, and algorithms/flows are listed on the corresponding pages and available for reproducibility. The code used for the tuning process (\texttt{HpTuning}), running meta-learning (\texttt{mtlSuite}), and performing the graphical analyses (\texttt{DecisionTreeTuningAnalysis}) are hosted at \texttt{GitHub}. These repositories are also listed in Table~\ref{tab:repo}. Instructions to run each project may be found directly at the correspondent websites.


\begin{table}[ht!]
\footnotesize
\centering
\caption{Repositories with tools developed by the authors and results generated by experiments.}
\begin{tabular}{ll}
        
    \toprule
    \textbf{Task/Experiment} & \textbf{Website/Repository} \\
    \midrule
    \rule{0pt}{3ex}
    Hyperparameter tuning code & \url{https://github.com/rgmantovani/HpTuning} \\
    \rule{0pt}{3ex}
    Hyperparameter tuning tasks & \url{https://www.openml.org/s/50} \\
    \rule{0pt}{3ex}
    Meta-learning code & \url{https://github.com/rgmantovani/mtlSuite} \\
    \rule{0pt}{3ex}
    Graphical Analysis & \url{https://github.com/rgmantovani/DecisionTreeTuningAnalysis} \\
    \bottomrule

\end{tabular}
\label{tab:repo}
\end{table}



\section{Hyperparameter tuning of decision trees}
\label{sec:results}

This section presents the results and analysis of optimizing the hyperparameters of \acrshort{dt} algorithms. These empirical findings aim to provide a comprehensive understanding of tuning the hyperparameter values for decision trees and offer guidance on the most effective techniques to perform this task while considering the criteria of improving predictive performance and minimizing computation cost.


\subsection{Is hyperparameter tuning necessary for decision trees?} 
\label{subsec:j48results}

\begin{figure*}[ht!]
    \includegraphics
    [width = \textwidth]
    {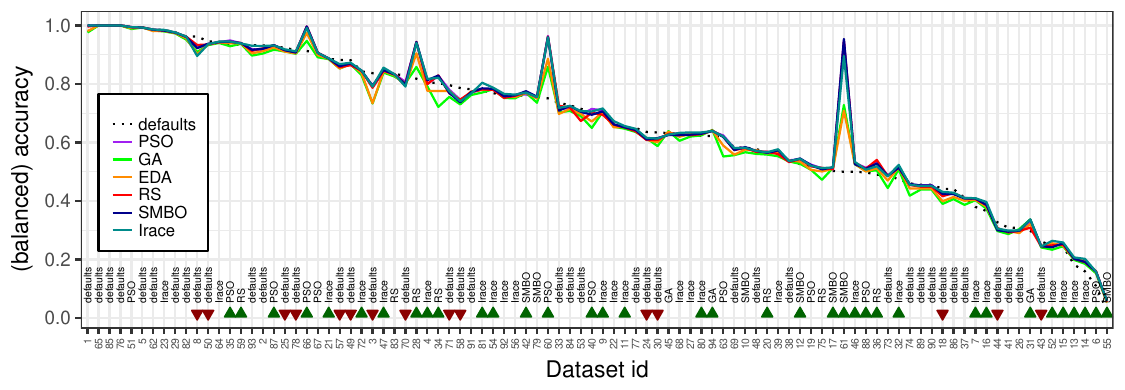}%
    \caption{Hyperparameter tuning results for the J48 algorithm.}
    \label{fig:j48_results}

    \vspace{0.5cm}
    \includegraphics
    [width = \textwidth]
    {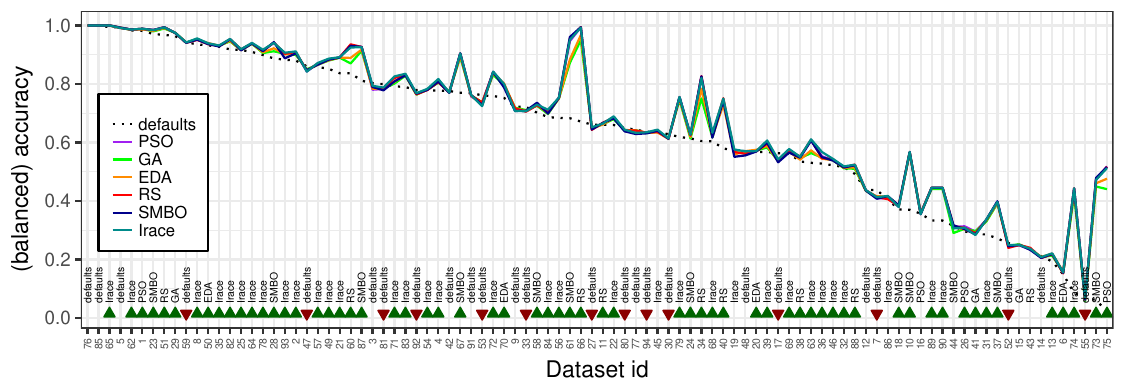}%
 \caption{Hyperparameter tuning results for the \acrshort{cart} algorithm.}
 \label{fig:cart_results}
\end{figure*}

Tuning results for J48 and \acrshort{cart} algorithms are depicted in Figure~\ref{fig:j48_results} and Figure \ref{fig:cart_results}, respectively. 
These figures show the predictive performance
\new{in terms of \acrshort{bac} values averaged over the $30$ repetitions (y-axis), for each tuning technique and default values over all datasets (x-axis) presented in decreasing order of the default results.} The name of the tuning technique that achieved the best predictive performance is shown above the x-axis \new{for each dataset}. The Wilcoxon paired test was applied to assess the statistical significance of the results obtained by this \textit{best} technique when compared to the results using default values. The test was applied to the solutions obtained from the $30$ repetitions (with $\alpha=0.05$). An upper green triangle 
(\textcolor{green}{$\blacktriangle$}) at x-axis identifies datasets where statistically significant improvements were detected after applying the hyperparameter tuning technique. On the other hand, every time a red down triangle~
(\textcolor{red}{$\blacktriangledown$}) is presented, the use of defaults was statistically better than the use of tuning techniques.

Tuning the \acrshort{hp} values of J48 (Fig.~\ref{fig:j48_results}) has not led to patent improvements in predictive performance for most of the datasets.
In general, the small peaks of improvements due to hyperparameter tuning show that default values are inappropriate for specific cases. This occurred, for instance, for the datasets with the ids = \{$36,\:46,\:61,\:88$\}. 
When the Wilcoxon statistical paired-test is applied, comparing defaults with the best tuning technique, they show that tuned trees were better overall than those with default values with statistical significance in $36/94$ ($\approx38\%$) datasets. In most of these situations, the \acrshort{irace}, \acrshort{pso}, or \acrshort{smbo} techniques produced the best results. Default values were significantly better in $15/94$ of the cases, and the remaining situations ($43/94$) did not present statistically significant differences (the approaches tied). 
Additionally, all tuning techniques presented similar performances, with few exceptions, since most curves overlap.  


On the other hand, the \acrshort{cart} algorithm benefits most by hyperparameter tuning. For \new{most} of the datasets analyzed, the use of tuned settings improved the predictive performance with statistical significance when compared with the use of default values in $62/94$ ($\approx66\%$) of the cases. 
Default values were better than tuned ones in $14/94$ ($\approx15\%$) of the cases, while there was no significant statistical in the remaining ($18/94$) datasets.
It must be observed that the \acrshort{irace} and \acrshort{smbo} were the best optimization techniques, regarding just the predictive performance of the induced models. 

\begin{figure*}[ht!]
    \subfloat[\new{J48 performances}]
    {\includegraphics
    [width=0.47\textwidth]
    {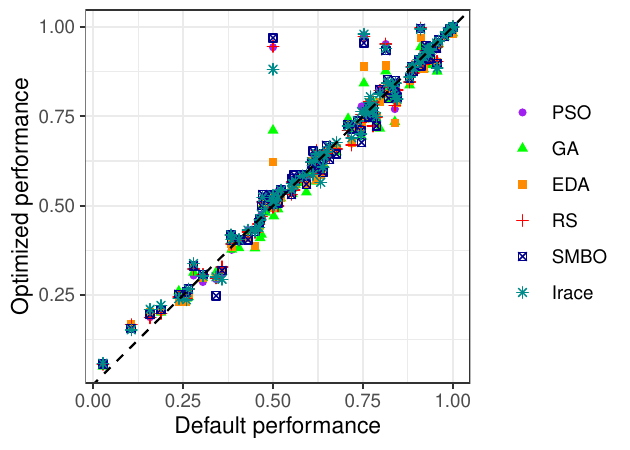}%
        \label{fig:scatter_j48}
    }
    \vspace{0.5cm}
    \subfloat[\new{\acrshort{cart} performances}]
    {\includegraphics
    [width=0.47\textwidth]
    {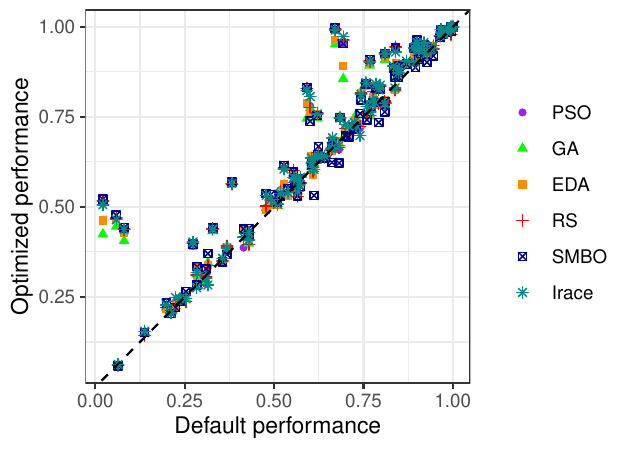}%
        \label{fig:scatter_cart}
    }
    \centering 
    \caption{\new{Scatter plot comparing performances obtained by default hyperparameter values against the optimized ones.}}
    \label{fig:scatter_performance}
\end{figure*}

\new{Figure~\ref{fig:scatter_performance} compares obtained performance of the default \acrshort{hp}  settings against the optimized ones for both \acrshort{dt} induction algorithms. The performance values obtained by the default \acrshort{hp}s are projected in the x-axis and the optimized values in the y-axis. Different shapes and colors represent different tuning techniques. The dotted black line represents the reference where defaults and optimized performances would be equal. Thus, points above the line indicate tuning provided improvements, while points below indicate worsening. Comparing both figures, we can see slightly different patterns. J48 points are mostly concentrated around the dotted black line (above and below). Some exceptions exist where tuning solutions obtained high improvements, especially for datasets with defaults obtained \acrshort{bac} values higher than 0.5. Conversely, \acrshort{cart} points are mostly above the black line, showing that tuning benefits a wider range of datasets.}


\begin{figure*}[ht!]
    \includegraphics 
    [width = \textwidth]
    {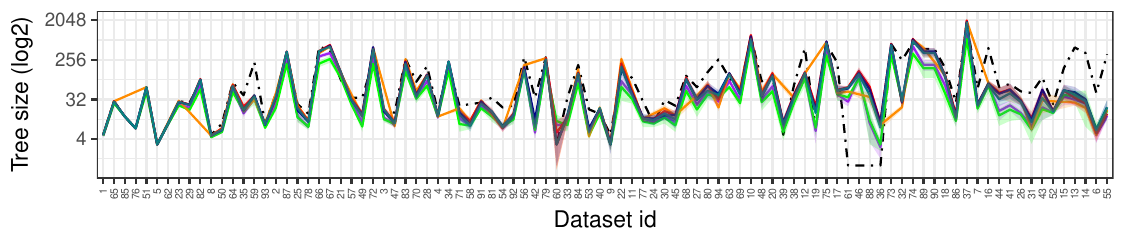}%
    \caption{J48 average tree size ($log_2$).}
    \label{fig:j48_tree_size}
    \vspace{0.5cm}
    \includegraphics 
    [width = \textwidth]
    {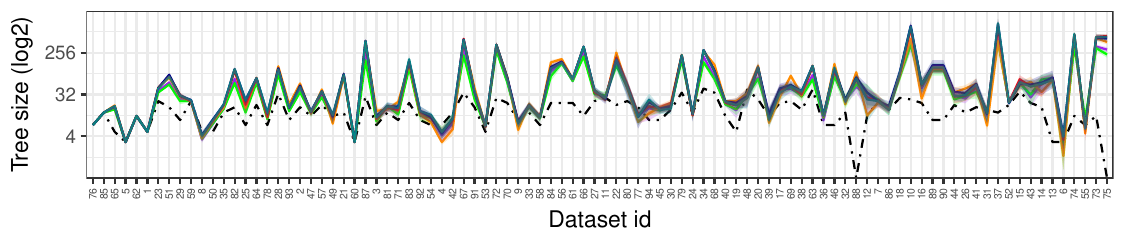}%
    \caption{CART average tree size ($log_2$).}
    \label{fig:cart_tree_size}
\end{figure*}

In addition to evaluating predictive performance, we also analyzed the effect of tuning techniques on tree size, which is given by the number of nodes in the final model. \new{It is important to mention that the tree's interpretability mostly depends on its size}. Thus, larger trees are usually more difficult to understand than smaller ones. The tree sizes for J48 and \acrshort{cart} algorithms are \new{shown} in Figure~\ref{fig:j48_tree_size} and Figure~\ref{fig:cart_tree_size}, respectively. The order of the datasets in these figures is the same as that of the models' \acrshort{bac}, \new{and line colors represent the same tuning techniques}.


Regarding the J48, in most cases, default values (dotted black line) induced trees larger than those obtained by the hyperparameters suggested by tuning techniques. 
For \new{most} of the multi-class problems (datasets most to the right \new{on} the charts), the tuned trees were also smaller than those induced using default values. 
\new{Conversely}, for the \acrshort{cart} algorithm, 
the trees induced with tuned \acrshort{hp} settings have similar or larger sizes than those induced by default values. 
The comparison among the tuning techniques showed results different from those obtained for the J48 algorithm. The tuning techniques led to the induction of \acrshortpl{dt} with similar sizes. However, the \acrshortpl{dt} induced when \acrshort{irace} was used were slightly larger and with better predictive performance than those induced using the other optimization techniques. 


\new{These results might reflect a biased design related to the origin of the \acrshort{dt} induction algorithms. \acrshort{cart}~\citep{Breiman:1984} was first developed by statisticians, who, at the time, had more interest in interpretable small but maybe underfitting trees. On the other hand, C4.5 (J48)~\citep{Quinlan:1993} was developed by computer scientists mainly concerned with maximum accuracy. Figures~\ref{fig:j48_results} to~\ref{fig:cart_tree_size} corroborate this bias, showing that default \acrshort{hp} values define different hyperparameter profiles for these algorithms: J48 trees tend to be deeper and more accurate at the same time \acrshort{cart} trees are shallower and not so accurate compared with their tuned versions.}


The~\acrshort{cart} hyperparameters' distributions found by the tuning techniques are presented in~Figure~\ref{fig:cart_distributions}. Again, unlike J48, \acrshort{cart} tuned trees were obtained from values substantially different from the default values. This is more evident for the numerical hyperparameters, as shown in Figures~\ref{fig:cart_distributions} (a) to (d). The `\texttt{cp}', `\texttt{minbucket}', and `\texttt{minsplit}' values tend to be smaller than default values. For `\texttt{maxdepth}', a wide range of values is tried, indicating a possible dependence on the input problem (dataset).
However, the categorical hyperparameters' distributions, shown in Figures~\ref{fig:cart_distributions}~(e) and (f), are very uniform, indicating that their choices may not influence the final predictive performance and could even act as ``noise'' when tuning is performed.


\subsection{Which hyperparameter tuning technique should I use?}
\label{subsec:stats}

Considering that hyperparameter tuning may improve predictive performance (mainly for CART) and/or \new{reduce} tree size (mainly for J48), the practitioner has to decide which tuning technique to choose. The previous analyses suggest that most of the techniques have a similar behavior. In order to compare them across all datasets, we applied the Friedman test~\citep{Demvsar:2006}, with significance levels at $\alpha = 0.05$ to evaluate the statistical significance of the experimental results. The null hypothesis states that all classifiers induced with the hyperparameter settings found by the tuning techniques and the classifier induced by default values are equivalent concerning predictive \acrshort{bac} performance. If the null hypothesis was rejected, the Nemenyi post-hoc test was applied, stating that the performances of two different techniques are significantly different if the corresponding average ranks differ by at least a \acrfull{cd}~value. 
Figure~\ref{fig:cd_diagram} presents \acrshort{cd} diagrams for the two \acrshort{dt} induction algorithms. Techniques are connected when there \new{are} \textit{no} statistically significant differences between them.


\begin{figure*}[h!t] 
\centering 
    \subfloat[J48 CD diagram with $\alpha=0.05$.]
    {
        \begin{tikzpicture}[xscale=2]
            \node (Label) at  (00.4184,0.7) {\tiny{CD}}; 
            \draw[decorate,decoration={snake,amplitude=.4mm,segment length=1.5mm,post length=0mm}, very thick, color = black](00.2857, 0.5) -- (00.5511, 0.5);
            \foreach \x in {00.2857,00.5511} \draw[thick,color = black] (\x, 0.4) -- (\x, 0.6);
            
            \draw[gray, thick](00.2857, 0) -- (02.0000, 0);
            \foreach \x in {00.2857,00.5714,00.8571,01.1429,01.4286,01.7143,02.0000}\draw (\x cm,1.5pt) -- (\x cm, -1.5pt);
            \node (Label) at (00.2857,0.2) {\tiny{1}};
            \node (Label) at (00.5714,0.2) {\tiny{2}};
            \node (Label) at (00.8571,0.2) {\tiny{3}};
            \node (Label) at (01.1429,0.2) {\tiny{4}};
            \node (Label) at (01.4286,0.2) {\tiny{5}};
            \node (Label) at (01.7143,0.2) {\tiny{6}};
            \node (Label) at (02.0000,0.2) {\tiny{7}};
            \draw[decorate,decoration={snake,amplitude=.4mm,segment length=1.5mm,post length=0mm}, very thick, color = black](00.7097,-00.2500) -- ( 01.0437,-00.2500);
            \draw[decorate,decoration={snake,amplitude=.4mm,segment length=1.5mm,post length=0mm}, very thick, color = black](00.9057,-00.5000) -- ( 01.1791,-00.5000);
            \node (Point) at (00.7597, 0){};  \node (Label) at (0.5,-00.6500){\scriptsize{Irace}}; \draw (Point) |- (Label);
            \node (Point) at (00.9557, 0){};  \node (Label) at (0.5,-00.9500){\scriptsize{PSO}}; \draw (Point) |- (Label);
            \node (Point) at (00.9909, 0){};  \node (Label) at (0.5,-01.2500){\scriptsize{defaults}}; \draw (Point) |- (Label);
            \node (Point) at (01.7369, 0){};  \node (Label) at (2.5,-00.6500){\scriptsize{GA}}; \draw (Point) |- (Label);
            \node (Point) at (01.4329, 0){};  \node (Label) at (2.5,-00.9500){\scriptsize{EDA}}; \draw (Point) |- (Label);
            \node (Point) at (01.1291, 0){};  \node (Label) at (2.5,-01.2500){\scriptsize{RS}}; \draw (Point) |- (Label);
            \node (Point) at (00.9937, 0){};  \node (Label) at (2.5,-01.5500){\scriptsize{SMBO}}; \draw (Point) |- (Label);
        \end{tikzpicture}
        \label{fig:j48_cd_005}
    }
      \subfloat[CART CD diagram with $\alpha=0.05$.]{
        \begin{tikzpicture}[xscale=2]
        
        \node (Label) at  (00.4184,0.7) {\tiny{CD}}; 
        \draw[decorate,decoration={snake,amplitude=.4mm,segment length=1.5mm,post length=0mm}, very thick, color = black](00.2857, 0.5) -- (00.5511, 0.5);
        \foreach \x in {00.2857,00.5511} \draw[thick,color = black] (\x, 0.4) -- (\x, 0.6);
        
        \draw[gray, thick](00.2857, 0) -- (02.0000, 0);
        \foreach \x in {00.2857,00.5714,00.8571,01.1429,01.4286,01.7143,02.0000}\draw (\x cm,1.5pt) -- (\x cm, -1.5pt);
        \node (Label) at (00.2857,0.2) {\tiny{1}};
        \node (Label) at (00.5714,0.2) {\tiny{2}};
        \node (Label) at (00.8571,0.2) {\tiny{3}};
        \node (Label) at (01.1429,0.2) {\tiny{4}};
        \node (Label) at (01.4286,0.2) {\tiny{5}};
        \node (Label) at (01.7143,0.2) {\tiny{6}};
        \node (Label) at (02.0000,0.2) {\tiny{7}};
        \draw[decorate,decoration={snake,amplitude=.4mm,segment length=1.5mm,post length=0mm}, very thick, color = black](00.6749,-00.2500) -- ( 01.0271,-00.2500);
        \draw[decorate,decoration={snake,amplitude=.4mm,segment length=1.5mm,post length=0mm}, very thick, color = black](00.9271,-00.5000) -- ( 01.2914,-00.5000);
        \draw[decorate,decoration={snake,amplitude=.4mm,segment length=1.5mm,post length=0mm}, very thick, color = black](01.0911,-00.7500) -- ( 01.4191,-00.7500);
        \draw[decorate,decoration={snake,amplitude=.4mm,segment length=1.5mm,post length=0mm}, very thick, color = black](01.1914,-00.2500) -- ( 01.5271,-00.2500);
        \node (Point) at (00.7249, 0){};  \node (Label) at (0.5,-01.0500){\scriptsize{Irace}}; \draw (Point) |- (Label);
        \node (Point) at (00.9771, 0){};  \node (Label) at (0.5,-01.3500){\scriptsize{RS}}; \draw (Point) |- (Label);
        \node (Point) at (01.0683, 0){};  \node (Label) at (0.5,-01.6500){\scriptsize{PSO}}; \draw (Point) |- (Label);
        \node (Point) at (01.4771, 0){};  \node (Label) at (2.5,-01.0500){\scriptsize{defaults}}; \draw (Point) |- (Label);
        \node (Point) at (01.3691, 0){};  \node (Label) at (2.5,-01.3500){\scriptsize{GA}}; \draw (Point) |- (Label);
        \node (Point) at (01.2414, 0){};  \node (Label) at (2.5,-01.6500){\scriptsize{EDA}}; \draw (Point) |- (Label);
        \node (Point) at (01.1411, 0){};  \node (Label) at (2.5,-01.9500){\scriptsize{SMBO}}; \draw (Point) |- (Label);
        
        \end{tikzpicture}
        \label{fig:cart_cd_005}
    }
    
    \caption{Comparison of the \acrshort{bac} values of the hyperparameter tuning techniques according to the Nemenyi test. Groups of techniques that are not significantly different are connected.}  
    \label{fig:cd_diagram}
\end{figure*}


Considering $\alpha=0.05$, Figure~\ref{fig:cd_diagram}(a) depicts the comparison for the J48 \acrshort{dt}. One may note that there \new{are no statistical} differences between the top three best techniques: \acrshort{irace}, \acrshort{pso}, and \acrshort{smbo}. 
\acrshort{eda} and \acrshort{ga} obtained statistically inferior performance. 
In addition, models induced with the \acrshort{hp} found by the best three techniques were not statistically better than
those induced by default values. This is in agreement with the analysis of the previous section, i.e., J48 does not generally benefit from performing \acrshort{hp} tuning.

For the \acrshort{cart} algorithm (Figure~\ref{fig:cd_diagram}(b)), the best-ranked technique over all datasets was \acrshort{irace}, followed by \acrshort{rs}, but with no \new{statistical} significance between them.  \acrshort{rs} has also no statistical difference with \acrshort{pso}, the third ranked technique here and the second for J48.
Finally, 
\acrshortpl{dt} induced with default hyperparameter values obtained the worst performance and are statistically comparable only with \acrshort{ga} and \acrshort{eda}.
It is worth mentioning that \acrshort{irace} was the best tuning technique for both algorithms. \new{At the same time}, the statistical test did not show significant differences between \acrshort{irace} and \acrshort{pso} (J48, \acrshort{ctree}), and between \acrshort{irace} and \acrshort{rs} (\acrshort{cart}), it is easy to see that \acrshort{irace} is the preferred technique, presenting the lowest averaging ranking. 

\begin{figure}[ph!tb]
        \centering
        \includegraphics
        [scale=0.8]{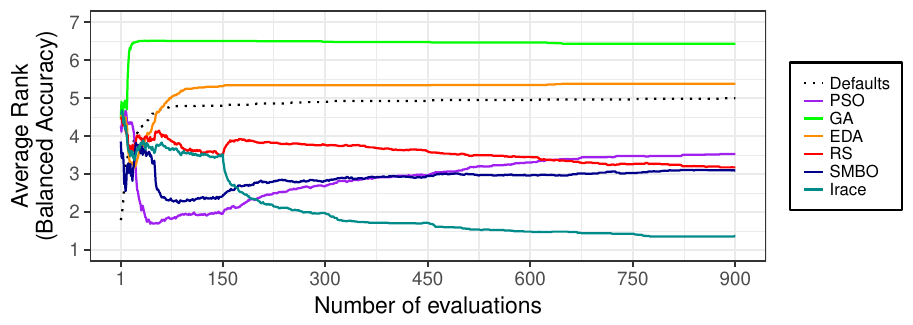}
        \caption{Average ranking curves of the tuning techniques for the J48 algorithm across all datasets.}
        \label{fig:j48_avg_rank}

    \vspace{0.5cm}
        \centering
        \includegraphics
          [scale=0.8]
        {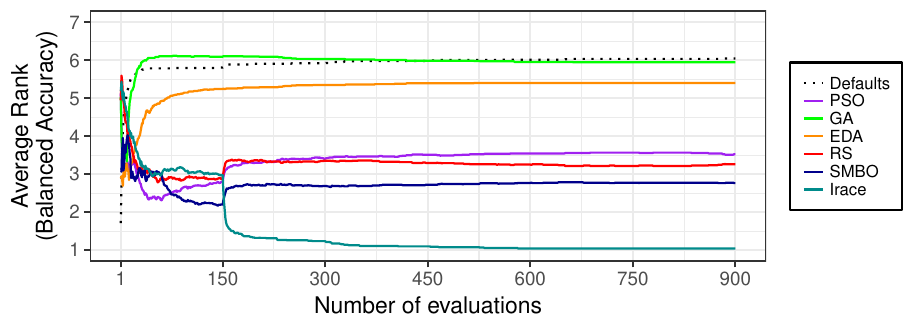}
        \caption{Average ranking curves of the tuning techniques for the CART algorithm across all datasets.}
        \label{fig:rpart_avg_rank}
\end{figure}

We can add another dimension to this analysis
examining the performance of various tuning techniques over time (number of evaluations). This is performed through their rankings, which are determined based on the predictive performance achieved using the best hyperparameter settings found by each technique within a specific number of evaluations.
Figure~\ref{fig:j48_avg_rank} shows the correspondent average rank (the smaller, the best) of each tuning technique over the evaluated budget (number of evaluations). Results are aggregated over all the $94$ datasets.


Using the average ranking curve (Figure~\ref{fig:j48_avg_rank}) a user may choose different techniques for tuning J48 according to the available budget size. For example, if tuning were performed with at most $b = 150$ evaluations, \acrshort{pso} and \acrshort{smbo} would be the best choices. With more than $200$ evaluations ($b > 200$), \acrshort{irace} surpasses all the techniques.


Similar behavior can be observed regarding the average rankings for \acrshort{cart}, i.e., different techniques are most suitable according to the budget size. If a tiny budget is provided ($b\leq50$), \acrshort{pso} is slightly better than the other techniques.  From $50<b\leq150$, the \acrshort{smbo} would be the best choice. 
On the other hand, with budgets ($b > 150$) \acrshort{irace} is the technique that best recommends values for \acrshort{cart} \acrshortpl{hp}.


\subsection{Which hyperparameters should be adjusted?}

Another approach to evaluate how the \acrshort{hp} are affecting the performance of the induced models when different tuning techniques are used is the use of~fANOVA (Functional ANOVA framework)\footnote{\url{https://github.com/automl/fanova}}, introduced in~\citet{Hutter:2014}. In that paper, the authors present a linear-time algorithm for computing marginal predictions and quantify the importance of single hyperparameters and interactions between them. The key idea is to generate regression trees that predict the performance of hyperparameter settings and apply the variance decomposition framework directly to the trees in these forests.

In the source article, the authors ran fANOVA with \acrshort{smbo} hyperparameter settings over some scenarios but never with more than $13.000$ hyperparameter settings. Here, a single execution of \acrshort{irace} generates $30 \times 10 \times 900 = 270.000$ evaluations. Thus, experiments using all techniques would have a high computational cost. Since \acrshort{irace} was the best technique overall for both algorithms, it was used to provide the \acrshort{hp} to this analysis. The experiments used settings from $3$ repetitions, and more memory was allocated to the fANOVA code.


\begin{figure*}[ht!]
    \flushright
    \subfloat[Functional ANOVA values for J48 hyperparameters.]
    {\includegraphics
    [width=0.95\textwidth]
    {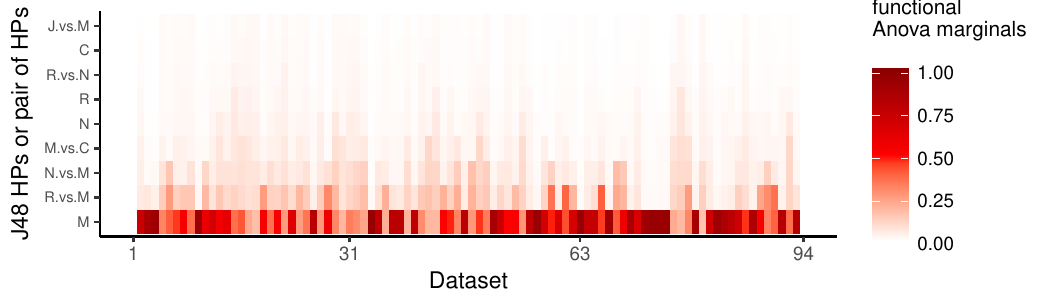}%
        \label{fig:fanova_j48}
    }
    \\
    \subfloat[Functional ANOVA values for \acrshort{cart} hyperparameters.]
    {\includegraphics
    [width=\textwidth]
    {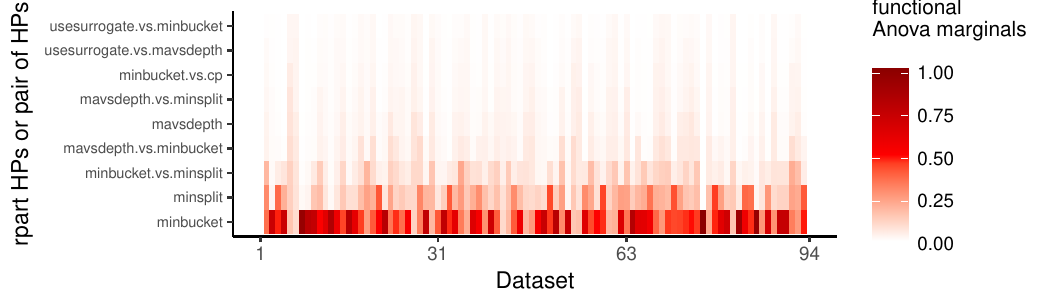}%
        \label{fig:fanova_cart}
    }
    \centering 
    \caption{Functional ANOVA hyperparameters marginal predictions for \acrshort{dt} algorithms regarding all dataset collection. Marginal predictions are scaled between zero and one.}
    \label{fig:fanova_params}
\end{figure*}


Figure~\ref{fig:fanova_params} shows the results for both \acrshort{dt} algorithms. In the figure, the x-axis shows all datasets while the y-axis presents the hyperparameters' importance regarding fANOVA. The larger the importance of a hyperparameter (or pair of them), the darker its corresponding square, i.e., the more important the hyperparameter is for inducing trees in the dataset (scaled between zero and one). 

In the figure, any hyperparameter (or their combination) whose contribution to the performance of the final models was lower than $0.005$ was removed. Applying this filter reduced the hyperparameters in focus, but even so, most of the rows in the heatmap are almost white (light red). This analysis shows that most combinations have little contribution to the performance of the induced \acrshortpl{dt}.

\begin{sloppypar}
In Figure~\ref{fig:fanova_params}(a), fANOVA indicates that the J48 performance was most influenced by the hyperparameter \texttt{M}, alone or in combination with another hyperparameter (\texttt{R}, \texttt{N} or \texttt{C}). 
The hyperparameter \texttt{M} defines the minimum number of instances in a leaf, influencing the size of the trees (models). The lower its value, the larger the trees and the better the models. Similar reasoning can be made with the \texttt{C} hyperparameter, which controls the pruning confidence of the pruning procedure. Depending on its value, the induction algorithm will prune the tree more or less, affecting the model's size. This corroborates the earlier discussion conducted in~Subsection~\ref{subsec:j48results}.

Finally, the last two \acrshortpl{hp} (\texttt{R}, \texttt{N}) corroborate this hypothesis: \texttt{R} is a \acrshort{hp} that enables/disables the \acrfull{rep}~\citep{Esposito:1999}. This post-pruning method finds the smallest version of the most accurate subtree concerning the pruning set. \texttt{N} is its conditional \acrshort{hp} that defines the number of inner folds used when post-pruning is enabled.


For \acrshort{cart}, we observed the same behavior, but conducted by the `\texttt{minbucket}' and `\texttt{minsplit}' \acrshortpl{hp}. \new{These \acrshortpl{hp}} are mainly responsible for the performance of the induced \acrshort{dt}s, as may be seen in Figure~\ref{fig:fanova_params}(b).
The former \acrshort{hp} is a counterpart of J48's  \texttt{M}, and affects the size of the trees. The latter determines when a split must occur: the lower its value, the larger the tree since more splits are required. Therefore, \texttt{minsplit} is also related to J48 findings regarding the \acrshort{hp} affecting the size of the tree.
However, unlike J48, the \acrshort{hp} values found by the tuning techniques for \acrshort{cart} are substantially different from the default values. These achievements unveil results discussed in the previous subsections, showing that a small subset of hyperparameters seems to influence the final performance of the induced \acrshortpl{dt} and \new{ the models' size}. The major difference is that J48 default settings can generate bigger trees as necessary to improve its predictive performance.
\end{sloppypar} 


\section{In which situations tuning of trees should be required?}
\label{sec:mtl_dt}

Considering that tuning the hyperparameters of \acrshort{dt} can significantly improve the predictive performance for some problems, a step forward is to investigate whether it is possible to predict for which problems the tuning should be performed and understand how to make informed decisions. \acrshort{mtl} and its ``\textit{inherent interpretability}'' could be explored for this aim.

In recent years, \acrfull{mtl}~\citep{Brazdil:2009} has been largely used for algorithm selection~\citep{Ali:2006}, algorithm ranking~\citep{Kanda:2016}, prediction of the performance of ML algorithms~\citep{Reif:2014}, and development of \acrshort{automl} tools~\citep{Feurer:2020, Gijsbers:2021}. \acrshort{mtl} learns from many previous experiences, i.e., applying different learning algorithms to many datasets, inducing a model capable of selecting the most promising algorithm for a new dataset.
\acrshort{mtl} has also been used to explain the effect of noise filtering techniques~\citep{Garcia:2019}, data imbalancement~\citep{Barella:2021}, and when performing \acrshort{hp} tuning~\citep{Sanders:2017, Mantovani:2019}. Therefore, we adapted the \acrshort{mtl} framework proposed by \citet{Mantovani:2019} to conduct experiments. The experimental setup is presented in Table~\ref{tab:mtl_setup} and explained as follows.


\subsection{Meta-learning framework}

\begin{figure}[h!]
    \centering
    \includegraphics[width=\textwidth]{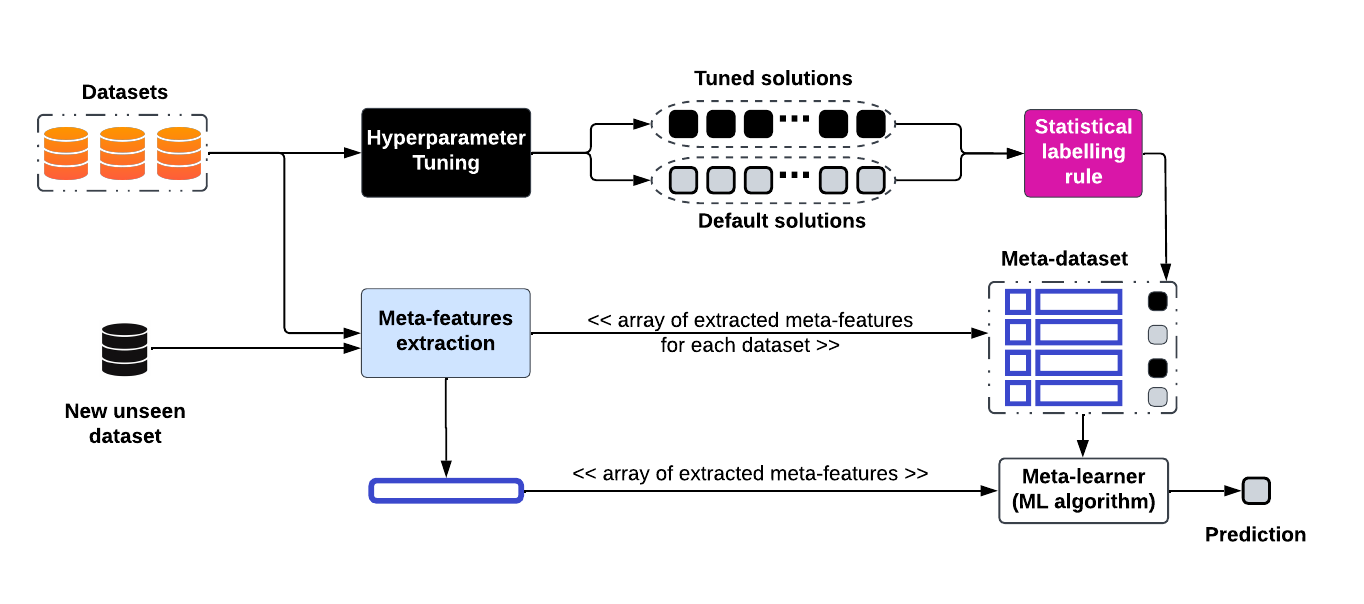}
    \caption{Meta-learning recommender system framework to predict when \acrshort{hp} tuning is required or not for \acrshortpl{dt}. Adapted from~\citet{Mantovani:2019}.}
    \label{fig:metalearning_framework}
\end{figure}

Figure~\ref{fig:metalearning_framework} depicts the general \acrshort{mtl} framework to identify whether \acrshort{hp} tuning for \acrshortpl{dt} is required or not.
\new{This framework has four sub-tasks:} i) hyperparameter tuning, ii) meta-feature extraction, iii) statistical labeling rule, and iv) meta-learning prediction. Each of them will be explained in sequence.


\subsubsection{Hyperparameter tuning task}

\acrshort{hp} tuning is performed across a collection of datasets. The output of this task is composed of two arrays: an array of the performances obtained through the tuning and an array of the performances obtained with default \acrshort{hp} values. Based on the results presented in Section~\ref{sec:results}, we enriched the ``\textit{meta-knowledge}'' required for the \acrshort{mtl} by performing additional hyperparameter tuning experiments. We selected $71$ additional datasets, totaling $165$ classification problems~\footnote{These additional datasets are indicated in Appendix~\ref{app:datasetsComplete}}. The additional \acrshort{hp} tuning jobs for J48 and \acrshort{cart} were performed with a budget size of $b = 900$ evaluations and, using the \acrshort{irace} technique, executed $30$ times with different seeds.
It is important to mention that tuning jobs were only required for the datasets not included in previous experiments. Datasets whose results were already available did not need to be tuned again.


\subsubsection{Meta-feature extraction}

The second required sub-task runs some data descriptors to extract likely relevant characteristics (\textit{meta-features}) from the datasets. These meta-features must be sufficient to describe the main aspects of the dataset necessary to distinguish the predictive performance obtained by defaults and tuned \acrshort{hp} settings.  Table~\ref{tab:meta-features} describes the main categories of the meta-features
implemented by the \texttt{pymfe} Python library~\citep{Alcobaca:2019}\footnote{A complete list of the \texttt{pymfe} available meta-features can be found here: \url{https://pymfe.readthedocs.io/en/latest/auto_pages/meta_features_description.html}.}.
Here, we explored the measures from simple/general (\textit{simple}) and data complexity (\textit{complex}) besides the combination of the $80$ measures from all categories presented in Table~\ref{tab:meta-features}.


\begin{table}[h!]
\footnotesize
\caption{Categories of meta-features used in meta-learning experiments}.
\label{tab:meta-features}
\centering
\begin{tabular}
{lrp{7cm}}
\toprule
\textbf{Category} & \textbf{\#} & \textbf{Description}  \\

\noalign{\smallskip}\hline\noalign{\smallskip}
\rule{0pt}{1ex} 
Simple/General & 17  & General measures about the input dataset, such as the number of (binary/categorical) attributes, examples and classes \\
\rule{0pt}{3ex} 
Statistical & 7  & Compute statistical descriptors about the input dataset, such as: minimum, maximum, mean and median values from each numerical attribute; skewness, kurtosis, correlation values, and so on. \\
\rule{0pt}{3ex} 
Information-theoretic & 8 & Information theory descriptors that measure: concentration of distinct attributes; entropy; mutual information; and noisiness of attributes. \\
\rule{0pt}{3ex} 
Model-based (trees) & 17 & A decision tree is applied to the dataset and statistics of nodes, leafs and branches are extracted.\\
\rule{0pt}{3ex} 
Landmarking & 8 & The predictive performance and runtime of simple \acrshort{ml} algorithms \\
\rule{0pt}{3ex} 
Data Complexity & 14 & Measures that analyze the complexity of a problem, such as the attributes values, the separability of the classes, and geometry/topological properties.
\\
\rule{0pt}{3ex} 
Complex Networks & 9 & A complex network (graph) is built with the dataset's instances and descriptors extracted from this graph: closeness and betweenness centralities, Hub score, average path length, and son on.\\
\midrule
\rule{0pt}{1ex} 
\textbf{Total} & 80 \\

\bottomrule

\end{tabular}
\end{table}


\subsubsection{Statistical labeling task}

In order to create each instance of the supervised problem at the meta-level, a label \new{must} be assigned to every dataset and combined with its respective meta-features. 
Hence, the third sub-task is a simple procedure that compares performance distributions through statistical tests. The Wilcoxon paired-test is applied with $\alpha=0.05$ to compare models' predictive performance using tuned and default \acrshort{hp} settings. Thus, given a dataset, if its tuned solution was significantly better than the default, its corresponding label assumes class 0 (`\texttt{Tuning}'), otherwise class 1 (`\texttt{Default}').

The labels yielded by this sub-task
are concatenated with their corresponding meta-features to create a \textit{meta-dataset}. Hence, each example (row) in the meta-dataset corresponds to a conventional dataset used in the tuning task and its corresponding label (`\texttt{Tuning}' or `\texttt{Default}'). Meta-features and labels compose the meta-knowledge of the \acrshort{mtl} recommendation problem.

Two meta-datasets were generated, one for each \acrshort{dt} algorithm. J48 meta-dataset \new{has} the \texttt{Default} class prevailing in two-thirds of the examples (datasets). \acrshort{cart} presents the opposite scenario, with two-thirds of examples prevailing in the \texttt{Tuning} class. Table~\ref{tab:meta_datasets} summarizes meta-datasets' main information, namely the number of meta-examples and meta-features, and the class distribution.


\begin{table}[h!]
\footnotesize
    \caption{Meta-datasets generated for meta-learning experiments and their class distributions using Wilcoxon paired-test with $\alpha=0.05$.}
    \label{tab:meta_datasets}
    \centering
    \color{black}{
    \begin{tabular}{lcccc}
    
    \toprule
    \multirow{2}{*}{\textbf{Meta-dataset}}  
    & \textbf{Meta} & \textbf{Meta} & \multicolumn{2}{c}{\textbf{Class Distribution}} \\
    & \textbf{examples} & \textbf{features} & \textit{Tuning} & \textit{Default} \\
    \midrule
    \rule{0pt}{1ex} 
    \multirow{1}{*}{J48} &  165 & 80 & 57 & 108 \\
    \multirow{1}{*}{\acrshort{cart}} & 165 & 80 & 111 & 54 \\
    \bottomrule
    \end{tabular}
    }
\end{table}

 
\subsubsection{Meta-learning predictions}

Finally, the fourth and last sub-task involves meta-learning training. A \acrshort{ml} algorithm, referred to as meta-learner, is trained using a meta-dataset to predict whether tuning is required.
We selected seven \acrshort{ml} algorithms that follow different learning biases so they explore the mapping between meta-features and labels differently. The following algorithms were evaluated as meta-learners: \acrfull{cart}, \acrfullpl{svm}, \acrfull{rf}, \acrfull{knn}, \acrfull{nb}, \acrfull{lr}, and \acrfullpl{gp}.

For comparison, it is important to mention that two baselines were also evaluated: Random and ZeroR. The former predicts a class by chance, while the latter predicts the majority class. 
Thus, it is expected that meta-models trained in the meta-datasets outperform them. Otherwise, no learning is happening.

These algorithms were trained in both meta-datasets using a $10$-fold \acrshort{cv} resampling strategy, repeated $10$ times with different seeds, and used their default \acrshort{hp} settings. Their predictions were assessed using the \acrfull{auc}, a more robust performance measure than \acrshort{bac} for binary classification problems\footnote{The \acrshort{bac} measure was preferred at the tuning level because data collection contains binary and multiclass classification problems.}. The complete \acrshort{mtl} setup is detailed in Table~\ref{tab:mtl_setup}.


\begin{table}[ht!]
\footnotesize
\centering
\caption{Meta-learning experimental setup.}
\begin{tabular}
    {@{\extracolsep{\fill}}lll}

\toprule
\textbf{Element} & \textbf{Method} & \textbf{Package (language)} \\
\midrule

  \rule{0pt}{1ex} 

\multirow{2}{*}{Meta-features} & All measures of features categories & pymfe (Python) \\
& described in Table~\ref{tab:meta-features} \\
\midrule

\multirow{3}{*}{Label definition} & Statistical labeling rule & \multirow{3}{*}{stats (R)} \\
& Wilcoxon paired-test & \\
& Significance level $\alpha = 0.05$ & \\
\midrule

\multirow{2}{*}{Meta-datasets} & J48 tuning recommendation & \multirow{2}{*}{-} \\
& CART tuning recommendation & \\
\midrule

\multirow{7}{*}{Meta-learner} & \acrfullpl{svm} & \texttt{e1071} (R) \\
 & \acrfull{cart} & \texttt{rpart} (R) \\
 & \acrfull{rf} & \texttt{randomForest} (R) \\
 & \acrfull{knn} & \texttt{kknn} (R) \\
 & \acrfull{nb} & \texttt{e1071} (R) \\ 
 & \acrfull{lr} & \texttt{gbm} (R) \\
 & \acrfullpl{gp} & \texttt{kernlab} (R) \\
\midrule

\multirow{2}{*}{Baselines} & Majority (ZeroR) & \multirow{2}{*}{mlr (R)} \\
& Random &  \\
\midrule
 
Resampling & $10$-fold CV & \texttt{mlr} (R) \\
 \midrule
 
\multirow{2}{*}{Repetitions} & $10$ times with different seeds & - \\
& seeds = $\{1, \ldots, 10\}$ & - \\
\midrule

Evaluation measures & AUC & \texttt{mlr} (R) \\
\bottomrule
 
\end{tabular}
\label{tab:mtl_setup}
\end{table}


\subsection{Meta-learning results}

\begin{figure}[htb]
    \centering
    \includegraphics[width=\textwidth]{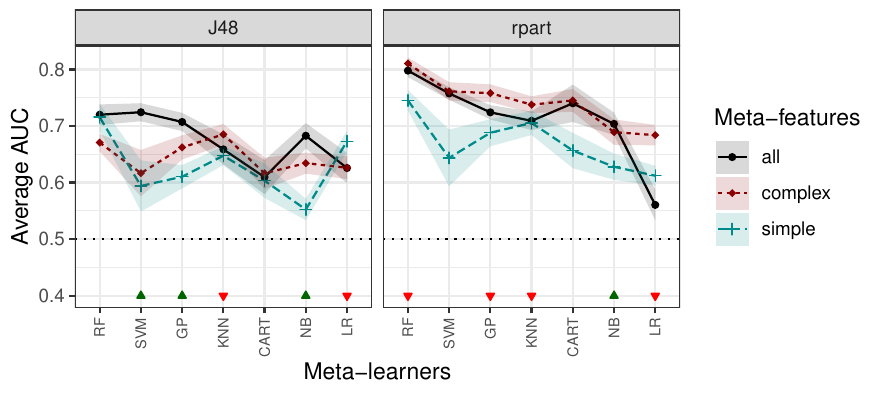}
    \caption{Meta-learners average AUC results on DTs meta-datasets. Black dotted lines at $AUC=0.5$ represent the predictive performance of ZeroR and Random meta-models.}
    \label{fig:mtl_dt_results}
\end{figure}

The predictive performance of the meta-learners when applied to the \acrshortpl{dt} meta-datasets are summarized in Figure~\ref{fig:mtl_dt_results}. All the meta-learners are listed in the x-axis, while their correspondent \acrshort{auc} values are projected the in y-axis. The colors of the lines represent the meta-features.

Results for these three sets of meta-features are presented using different colors and line types to distinguish between them.
The Wilcoxon paired test with a significance level $\alpha=0.05$ was applied to assess the statistical significance of the meta-models' performance obtained by the top two best meta-datasets regarding the meta-feature sets. These differences are indicated in the x-axis. Green triangles identify cases where using ``all'' the meta-features yielded statistically better results. On the other hand, red triangles indicate where one of the other meta-feature sets (``simple'' or ``complex'') was significantly better. For the remaining cases, the predictive performance values of the meta-models were equivalent.
The baselines (Random and ZeroR) obtained \acrshort{auc} of $0.5$ in all meta-datasets, represented by the dotted line at $AUC=0.5$ in Figure~\ref{fig:mtl_dt_results}. 

The best results considering J48 were obtained by the \new{\acrshort{rf}},\:\acrshort{svm}, and\:\acrshort{gp}\ meta-learners. 
They obtained their best predictive performances using all the available meta-features (``all'' set of meta-features). 
For the \acrshort{rf} meta-learner, using only the simple meta-features also generated accurate meta-models with \acrshort{auc} $>0.7$. 
However, the complete meta-feature set generally provided the best results for most algorithms.
In addition, these meta-models achieved \acrshort{auc} values $\in [0.67, 0.74]$, which are noticeably better than predictions based on chance. 


For the  \acrshort{cart} algorithm, the best results were also obtained by \acrshort{rf} meta-models, but using data complexity meta-features (``complex''). However, there is no significant performance difference to the complete set of meta-features (``all''). 
These meta-models achieved \acrshort{auc} values near $0.8$.  
These two approaches (``complex'' and ``all'') performed quite similarly when explored by the meta-learners. 
In addition to \acrshort{rf}, all the algorithms, except for \acrshort{nb} and \acrshort{lr}, induced meta-models with predictive performance $AUC > 0.75$. It is also noteworthy that the meta-model \acrshort{cart} is among the top three meta-learners for the \acrshort{cart} algorithm, while this meta-model was one of the worst regarding the J48 algorithm. 

Considering these two scenarios, \acrshort{mtl} results suggest that it is possible to predict whether tuning is required for \acrshortpl{dt}. Overall, the best results for J48 and \acrshort{cart}  were obtained when inducing RF meta-models with all the available meta-features (``all''). 
Thus, the following analyses for these meta-datasets were performed using the complete set of meta-features.


\subsection{Meta-models' interpretability}
\label{subsec:explainable}

\begin{figure}[htb!]
    \centering
     \begin{tabular}{cc}
        \subfloat[J48]{
        \includegraphics
         [width = 0.8\textwidth]
        {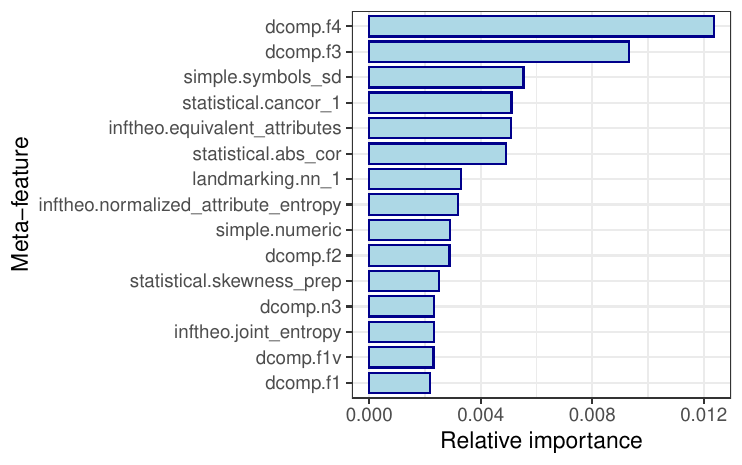}} 
        \\
        \subfloat[\acrshort{cart}]{
        \includegraphics
         [width = 0.8\textwidth]
        {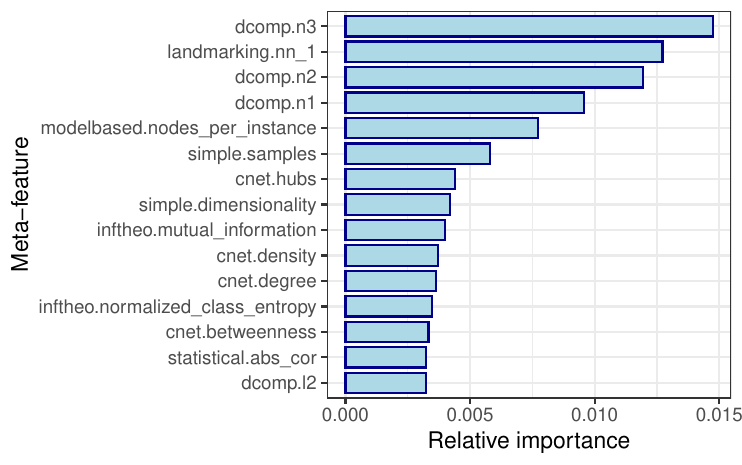}}       
    \end{tabular}
    \caption{Relative importance of the meta-features extracted from \acrshort{rf} meta-models.}
    \label{fig:rf-mfeat_dts}
\end{figure}

Since \acrshort{rf} meta-models \new{accurately} predicted the tuning necessity of \acrshortpl{dt}, we can unveil their predictions exploring the relative importance of the meta-features based on the Gini impurity index. Figure~\ref{fig:rf-mfeat_dts} shows the average relative importance values for top-15 meta-features in each \acrshort{dt} meta-dataset. The y-axis lists meta-features, 
from best to worst, according to their relative importance, projected in the x-axis.  The importance values for the experiments considering all meta-features are shown. 


As expected, the figure shows that different meta-features are important for different problems. Some of them contribute to both tasks, such as the absolute correlation of the attributes (\texttt{statistical.abs\_cor}) meta-feature, which is the 6th ranked meta-feature in J48 tuning recommendation and the 14th for \acrshort{cart}. 
However, in most cases, the most important meta-features are specific for each problem.

The most crucial meta-feature for the J48 problem was the data complexity measure ``\texttt{f4}''. 
This feature describes the collective attribute efficiency in a dataset. The second most important was ``\texttt{f3}'', which describes the maximum individual attribute efficiency.
The two data complexity features measure the discriminative power of the dataset's attributes.
The top-3 is completed with a simple measure, ``\texttt{symbols\_sd}'', which measures the standard deviation of categorical features presented in the dataset.
These most important metrics suggest that if a dataset has representative attributes, default \acrshort{hp} values are robust enough to solve it. Otherwise, \acrshort{hp} tuning is recommended.

For the \acrshort{cart} meta-dataset, the two most important 
meta-features are: ``\texttt{n3}'' and ``\texttt{nn}''. The first is a data complexity meta-feature, and the second is a traditional landmarking. Both measures estimate the error rate of the 1-NN classifier by leave-one-out or cross-validation. They measure how close the examples of different classes are. Low values mean that there is a large gap in the class boundary. 
The third most important is a data complexity meta-feature: ``\texttt{n2}'', the average intra-class and inter-class distances ratio used by a \acrshort{knn} algorithm to classify data examples. Low values indicate that examples from the same class lay closely in the feature space. 
This suggests that if the dataset's examples from the same class are close to each other, the 1-NN will find them closely in the feature space, and \acrshort{cart} would also be capable of solving it without tuning. On the other hand, examples would be more overlapped, and the \acrshort{hp} tuning would better fit the algorithm's learning bias.

In general, different aspects are being considered when choosing between tuned of default \acrshort{hp} settings for the target algorithms: the discriminative power of the attributes is important to recommend the J48 tuning, and the dispersion of datasets' examples from different classes to recommend \acrshort{cart} tuning. At least for both algorithms, it is important to verify the importance of the dataset's attributes before doing the \acrshort{hp} tuning, which makes sense for a \acrshort{dt} induction algorithm.


\section{Threats to Validity} 
\label{sec:validity}

In an empirical study design, methodological choices may impact the results obtained in the experiments. Next, the threats that may impact the results of this study are discussed.


\subsection{Construct validity}


The datasets used in the experiments were selected to cover a wide range of classification tasks with different characteristics. They were used in their original versions, i.e., no preprocessing was required since \acrshortpl{dt} can handle any missing information or data from different types. The only restriction adopted ensures that all classes in the datasets must have at least $10$ observations. Thus, stratification with $10$ outer folds can be applied. Of course, other datasets may be used to expand data collection if they obey the `stratified' criterion. However, the authors believe that adding datasets will not substantially change the overall tuning behavior of the algorithms investigated.


Regarding the \acrshort{dt} induction algorithms, \acrshort{cart} and J48 are among the most popular algorithms used in data mining~\citep{Wu:2009}. Experiments were focused on these algorithms due to the interpretability of their induced models and widespread use. They generate simple models, are robust for specific domains, and allow non-expert users to understand how the classification decision is made. The same experimental methodology and analyses can be applied to any other \acrshort{ml} algorithm.


Since a wide variety of datasets compose the data collection, some may be imbalanced. Thus, the \acrshort{bac} performance measure~\citep{Brodersen:2010} was used as a fitness function during the optimization process. Therefore, class distributions are being considered when assessing a candidate solution. The same performance measure is used to evaluate the final solutions returned by the tuning techniques. Other predictive performance measures can generate different results, depending on how they deal with data imbalance.


\new{The experimental methodology described in Section~\ref{sec:methods} considers the main paradigms for \acrshort{hp} tuning (black-box optimization) found in the literature~\citep{Sureka:2008, Sun:2013, Kotthoff:2016}, such that: a swarm method (\acrshort{pso}), an evolutionary method (\acrshort{ga}), the standard baseline of random search (\acrshort{rs}), and a state-of-the-art Bayesian technique (\acrshort{smbo}). 
Additionally, we included \acrshort{eda} and \acrshort{irace} because they have been explored recently for \acrshort{hp} tuning of other \acrshort{ml} algorithms, like \acrshortpl{svm}~\citep{Miranda:2014,Padierna:2017}, and we had already evaluated them in our previous paper~\citep{Mantovani:2019}.
They are ''non-conventional'' choices and can be considered novelties in decision tree \acrshort{hp} tuning. Hyperband and \acrshort{bohb} would fit in the two different \acrshort{hp} tuning paradigms used in the experiments, such as \acrshort{smbo} and \acrshort{rs}, since i) Hyperband focuses on speeding up \acrfull{rs} ''by formulating hyperparameter optimization as a pure-exploration adaptive resource allocation problem addressing how to allocate resources among randomly sampled hyperparameter configurations''~\citep{hyperband:2018}, while ii) \acrshort{bohb} is based on the combination of \acrshort{smbo} and Hyperband.
It is important to mention that the results reported in Subsection 5.1 are conditioned to the pool of tuning techniques selected in the experimental methodology. Thus, adding more tuning techniques can slightly change the best technique for some of the datasets.}


\subsection{Internal validity}


\begin{sloppypar}
\citet{Krstajic:2014} compared different resampling strategies for selecting and assessing the predictive performance of regression/classification models induced by \acrshort{ml} algorithms. In~\citet{Cawley:2010} the authors also discuss the overfitting in the evaluation methodologies when assessing \acrshort{ml} algorithms. They describe a so-called ``\textit{unbiased performance evaluation methodology}'', which correctly accounts for any overfitting that may occur in the model selection. The internal protocol described by the authors performs the model's selection independently within each fold of the resampling procedure. In fact, most of the current studies on hyperparameter tuning have adopted nested-\acrshortpl{cv}, including important autoML tools, like \texttt{Auto-WEKA}\footnote{\url{http://www.cs.ubc.ca/labs/beta/Projects/autoweka/}}~\citep{Thornton:2013,Kotthoff:2016} and \texttt{Auto-skLearn}\footnote{\url{https://github.com/automl/auto-sklearn}}~\cite{Feurer:2015B}. Since this paper aims to assess \acrshort{dt} induction algorithms optimized by hyperparameter tuning techniques, the nested \acrshort{cv} methodology is the best choice and was adopted in the experiments. 
\end{sloppypar}


In the experiments carried out for this study, all the default settings provided by the implementations of the tuning techniques were used. In fact, most of these default values have been evaluated in benchmark studies and reported to provide good predictive performance~\citep{Perez:2014}, while others (like \acrshort{pso}'s) showed to be robust in a high number of datasets. For \acrshort{eda} and \acrshort{ga}, there is no standard choice for their parameter values~\citep{Mills:2015}, and even adapting both to handle our mixed hyperparameter spaces properly, they performed poorly.
It suggests that a fine-tuning of their parameters would be needed. 
Since this would considerably increase the cost of experiments by adding a new tuning level (\textit{the tuning of tuning techniques}), and most of the techniques performed well with default values, this additional tuning was not assessed in this study.


Using a larger budget, with $5000$ evaluations for \acrshort{dt} tuning, was investigated in~\citet{Mantovani:2016}. The experimental results suggested that all the considered techniques required only $900$ evaluations to converge. The convergence here means the tuning techniques could not improve their predictive performance more than $x=10^5$ until the budget was consumed.
Actually, in most cases, the tuning reached its maximum performance after $300$ steps. Thus, a budget size of $900$ evaluations was therefore deemed sufficient. Results obtained with this budget value showed that the exploration in hyperparameter spaces led to statistically significant improvements in most cases.


\subsection{External validity}


Section~\ref{subsec:stats} presented statistical comparisons between tuning techniques. In~\citet{Demvsar:2006}, Dem\v{s}ar discusses the issue of statistical tests for comparisons of several techniques on multiple datasets reviewing several statistical methodologies. The method proposed as more suitable is the non-parametric analog version of ANOVA, i.e., the Friedman test, along with the corresponding Nemenyi post-hoc test. The Friedman test ranks all the methods separately for each dataset and uses the average ranks to test whether all techniques are equivalent. In case of differences, the Nemenyi test performs all the pairwise comparisons between the techniques and identifies significant differences. Thus, the Friedman ranking test followed by the Nemenyi post-hoc test was used to evaluate experimental results from this study.


The budget size adopted can directly influence the performance of the meta-heuristics, especially \acrshort{ga} and \acrshort{eda}. In~\citet{Hauschild:2011}, the authors recommend to use at least $100$ individuals to build a reliable \acrshort{eda} model, a suggestion followed in~\citet{Mantovani:2016}. In this extended version, the budget size was reduced, supported by prior analyses and tuning techniques adapted to work with the reduced number of evaluations. Increasing the population size would also increase both the number of iterations and the budget size. However, it has already been experimentally shown that just a small number of evaluations provide good predictive performance values~\citep{Mantovani:2016}. It is important to highlight that even using a small population the \acrshort{pso} technique reached robust results in a wide variety of tasks considering the three \acrshort{dt} algorithms investigated. At this point, the poor performance values obtained by \acrshort{ga} and \acrshort{eda} can be considered a limitation: they do not search properly for space under this budget restriction.


\section{Conclusions} 
\label{sec:conclusions}

This paper investigated the effects of the hyperparameter tuning profile of \acrshort{dt} induction algorithms. For this purpose, two of the most popular \acrshort{dt} algorithms were chosen as study cases: J48 and \acrshort{cart}.
An experimental analysis regarding the sensitivity of their hyperparameters was also presented. Experiments were carried out with $94$ public \texttt{\acrshort{openml}} datasets and six different tuning techniques. The performances of \acrshort{dt} induced using these techniques were also compared with \acrshortpl{dt} generated with the default hyperparameter values (provided by the correspondent R packages). The main findings are summarized below. 


\subsection{Tuning of J48}
\label{sub:j48_concl}

In general, hyperparameter tuning for J48 produced modest improvements when compared to the \texttt{RWeka} default values: the trees induced with tuned hyperparameter settings reached performances similar to those obtained by defaults. Statistically significant improvements were detected in only one-third of the datasets, often those datasets where the default values produced very shallow trees.

The J48 boolean hyperparameters are responsible for enabling/disabling some data transformation processes. In default settings, all of these hyperparameters are disabled. Thus, enabling them requires more time to induce and assess trees. Furthermore, the relative hyperparameter importance results (via fANOVA analysis) showed that these boolean hyperparameters are irrelevant for most datasets. Only a subset of hyperparameters (\texttt{R}, \texttt{C}, \texttt{N}, \texttt{M}) contributes actively to the performance of the final \acrshortpl{dt}.

Most of the related studies that performed some tuning for J48 tried different values for the complexity parameter (\texttt{C}). However, none tried hyperparameter tuning using reduced error pruning: enabling `\texttt{R}' and changing `\texttt{N}' values. 
The use of `\texttt{R}' and `\texttt{N}' options
may be a solution when tuning only `\texttt{C}' does not sufficiently improve performance. 

None of the related work used the \acrshort{irace} technique: they focused on \acrshort{smbo}, \acrshort{pso}, or another tuning technique. \acrshort{smbo} is often used with an early stopping criterion (a budget) since it is the slowest technique. However, it typically converged after relatively few iterations. If it is desirable to obtain good solutions faster, \acrshort{pso} might be recommended. However, for the J48 algorithm, the best technique concerning performance is \acrshort{irace}: it was better ranked, evaluated more candidates, and did not consume much runtime.

The default hyperparameter values of J48 proved effective for many datasets. This outcome can be because the default settings used by \texttt{RWeka} were chosen to aim to maximize the predictive performance on the \acrshort{uci} \acrshort{ml} repository datasets~\citep{Bache:2013}.


\subsection{Tuning of CART}
\label{sec:cart_concl}

Surprisingly, \acrshort{cart} was much more sensitive to hyperparameter tuning than J48. Statistically significant improvements were reached in two-thirds of the datasets, most of them with a high-performance gain.
Most hyperparameters control the number of instances in nodes/leaves used for splitting. These hyperparameters directly affect the size and depth of the trees. The experimental analyses showed that default settings induced shallow and small trees for most of the problems. These trees did not obtain good predictive performances. Where the defaults did grow large trees, the performance was similar to the optimized setting. In general, \acrshort{cart}'s default hyperparameter values induced trees, which are, on average, smaller than those produced by J48 under default settings. One reason that may also explain the poor \acrshort{cart}'s default performances would be the case that J48 hyperparameters were pre-tuned on \acrshort{uci} datasets while the \acrshort{cart} ones were not.

Our relative importance analysis indicated that hyperparameters such as `\texttt{minsplit}' and `\texttt{minbucket}' are the most responsible for the performance of the final trees. In the related literature, just two of the five works investigated the tuning of both. Even so, they used \acrshort{rs} and \acrshort{smbo} as tuning techniques. Experiments showed that for \acrshort{cart} hyperparameter tuning, the \acrshort{irace} technique significantly outperformed all the other ones (especially with $\alpha=0.1$). It evaluated a higher number of candidates during the search, and its running time was comparable to that of the meta-heuristics. Thus, \acrshort{irace} would be a good choice and might be further explored in future research.


\subsection{General scenario} 

Overall, the general picture showed us that tuning techniques can significantly improve the predictive performance of the \acrshortpl{dt}. Depending on the dataset, tuning techniques' performance values can be very small compared to those obtained by default HP settings. Hence, the results indicate that it is better to use the default settings for some optimization problems. 
\new{Different techniques are more suitable for different budget sizes when comparing tuning techniques.} If the user has a large enough budget (time or evaluations), \acrshort{irace} is a good choice. On the other hand, \acrshort{pso} and \acrshort{smbo} are the recommended techniques with faster convergence and faster results.

The fANOVA analysis also indicated that few of the HPs are effectively responsible for the predictive performance of the final trees. Similar results with different ML algorithms were reported in~\citet{Rijn:2017}.  
In this sense, the fANOVA framework is a powerful tool to reduce the search HP space and time spent with optimization. The last tuning experiment showed that a higher number of statistically significant improvements were obtained when a reduced HP space was used to tune both algorithms. 


\subsection{When to tune}

Lastly, we explored \acrshort{mtl} to unveil the tuning process so \acrshort{mtl} could explain when to use the tuning approach.
We observed that hyperparameter tuning provides the best results for datasets with many classes ($cls>8$), and when there are non-linear decision boundaries. On the other hand, defaults are adequate for simple classification problems, where there is a higher separability between the classes. It can be assumed that the more complex (difficult to classify) a dataset is, the more a \acrshort{dt} algorithm will benefit from hyperparameter tuning. 
 
Considering the algorithms investigated in this study, each presented a different behavior under tuning. 
Generally, it was possible to observe that the default hyperparameter values are suitable for many datasets. However, a fixed value would only suit some data classification tasks. It justifies and motivates the development of recommender systems able to suggest the most appropriate hyperparameter setting for a new problem.


\subsection{Future Work}

Our findings also point out to some future research directions. The data complexity characteristics provided \new{helpful} insight regarding which situations tuning or defaults should be used. However, it would be possible to make more accurate suggestions by exploring more concepts from the meta-learning field.

It would obviously also be interesting to explore other ML algorithms and their hyperparameters: not only \acrshortpl{dt} induction algorithms but many classifiers from different learning paradigms. The code developed in this study, which is publicly available, is easily extendable and may be adapted to cover a wider range of algorithms. The same can be said for the analysis.

All collected hyperparameter information might be leveraged in a recommendation framework to suggest hyperparameter settings. When integrated with \acrshort{openml}, this framework could have a great scientific (and societal) impact. The authors have already begun work in this direction.

\section*{Acknowledgements}

\begin{sloppypar}
The authors would like to thank the Coordena\c{c}\~{q}o de Aperfei\c{c}oamento de Pessoal de N\'{i}vel Superior (CAPES) for the financial support, the Brazilian National Council for Scientific and Technological Development (CNPq) for the grant \#409371/2021-1 (CNPq/MCTI/FNDCT No 18/2021), and specially to the grants \#2012/23114-9, \#2013/07375-0 and \#2015/03986-0 from S{\~a}o Paulo Research Foundation (FAPESP). EFOP-3.6.3-VEKOP-16-2017-00001: Talent Management in Autonomous Vehicle Control Technologies -- The Project is supported by the Hungarian Government and co-financed by the European Social Fund.
\end{sloppypar}

\section*{Declarations}

\textbf{Competing interests:} The authors have no competing interests to declare that are relevant to the content of this article.



\begin{appendices}

\newpage
\section{List of abbreviations used in the paper} 
\label{app:glossary}

\printnoidxglossary[type=\acronymtype, title=\empty]

\newpage
\clearpage
\section{List of OpenML datasets used in experiments} 
\label{app:datasetsComplete}

This appendix presents the full table of datasets used in both tuning and meta-learning experiments performed in this paper. For each dataset it is shown: the OpenML dataset name and id, the number of attributes (D), the number of examples (N), the number of classes (C), the number of examples belonging to the majority and minority classes (nMaj, nMin), the proportion between them (P), and whether the dataset was added to the enrichment step for meta-learning.

   
\begin{table*}[h!]
\centering
\footnotesize

\setlength{\tabcolsep}{3pt}
 \caption{(Multi-class) classification OpenML datasets (1 to 29) used in experiments. For each dataset it is shown: the OpenML dataset name and id, the number of attributes (D), the number of examples (N), the number of classes (C), the number of examples belonging to the majority and minority classes (nMaj, nMin), the proportion between them (P), and whether the dataset was added to the enrichment step for meta-learning.}
 \label{app:data1}

\begin{tabular}{clcrrrrrrc}
    \toprule   
   \textbf{Nro } & \textbf{ OpenML name} & \textbf{OpenML did} & \textbf{ D } & \textbf{ N } & \textbf{ C } & \textbf{ nMaj } & \textbf{ nMin } & \textbf{ P} & \textbf{Meta} \\ 
  \midrule

  1 & kr-vs-kp &   3 &  37 & 3196 &   2 & 1669 & 1527 & 0.91 & base \\ 
  2 & balance-scale &  11 &   5 & 625 &   3 & 288 &  49 & 0.17 & base\\
  3 & breast-cancer &  13 &  10 & 286 &   2 & 201 &  85 & 0.42 & $\bullet$ \\ 
  4 & mfeat-fourier &  14 &  77 & 2000 &  10 & 200 & 200 & 1.00 & base \\ 
  5 & breast-w &  15 &  10 & 699 &   2 & 458 & 241 & 0.53 & base \\ 
  6 & mfeat-karhunen &  16 &  65 & 2000 &  10 & 200 & 200 & 1.00 & $\bullet$ \\ 
  7 & mfeat-morphological &  18 &   7 & 2000 &  10 & 200 & 200 & 1.00& $\bullet$ \\ 
  8 & car &  21 &   7 & 1728 &   4 & 1210 &  65 & 0.05 & base \\ 
  9 & mfeat-zernike &  22 &  48 & 2000 &  10 & 200 & 200 & 1.00 & $\bullet$ \\ 
  10 & colic &  25 &  28 & 368 &   2 & 232 & 136 & 0.59 & base \\ 
  11 & optdigits &  28 &  65 & 5620 &  10 & 572 & 554 & 0.97  & base \\ 
  12 & credit-g &  31 &  21 & 1000 &   2 & 700 & 300 & 0.43 & base \\ 
  13 & pendigits &  32 &  17 & 10992 &  10 & 1144 & 1055 & 0.92 & $\bullet$ \\ 
  14 & dermatology &  35 &  35 & 366 &   6 & 112 &  20 & 0.18 & base \\ 
  15 & segment &  36 &  20 & 2310 &   7 & 330 & 330 & 1.00  & base \\ 
  16 & diabetes &  37 &   9 & 768 &   2 & 500 & 268 & 0.54 & $\bullet$\\ 
  17 & sick &  38 &  30 & 3772 &   2 & 3541 & 231 & 0.07 & $\bullet$ \\ 
  18 & sonar &  40 &  61 & 208 &   2 & 111 &  97 & 0.87 & base \\
  19 & haberman &  43 &   4 & 306 &   2 & 225 &  81 & 0.36 & base \\
  20 & spambase &  44 &  58 & 4601 &   2 & 2788 & 1813 & 0.65 & $\bullet$\\ 
  21 & tae &  48 &   6 & 151 &   3 &  52 &  49 & 0.94 & base  \\ 
  22 & heart-c &  49 &  14 & 303 &   5 & 165 &   0 & 0.00 & base\\ 
  23 & tic-tac-toe &  50 &  10 & 958 &   2 & 626 & 332 & 0.53 & $\bullet$\\ 
  24 & heart-statlog &  53 &  14 & 270 &   2 & 150 & 120 & 0.80 & base \\ 
  25 & vehicle &  54 &  19 & 846 &   4 & 218 & 199 & 0.91 & base \\ 
  26 & hepatitis &  55 &  20 & 155 &   2 & 123 &  32 & 0.26 & base \\
  27 & vote &  56 &  17 & 435 &   2 & 267 & 168 & 0.63 & $\bullet$\\ 
  28 & ionosphere &  59 &  35 & 351 &   2 & 225 & 126 & 0.56 & base \\ 
  29 & waveform-5000 &  60 &  41 & 5000 &   3 & 1692 & 1653 & 0.98 & $\bullet$ \\ 
 \bottomrule

\end{tabular}
\end{table*}


\begin{table*}[h!]
\centering
\footnotesize

\setlength{\tabcolsep}{3pt}
 \caption{(Multi-class) classification OpenML datasets (30 to 67) used in experiments. For each dataset it is shown: the OpenML dataset name and id, the number of attributes (D), the number of examples (N), the number of classes (C), the number of examples belonging to the majority and minority classes (nMaj, nMin), the proportion between them (P), and whether the dataset was added to the enrichment step for meta-learning.}
 \label{app:data2}
\begin{tabular}{clcrrrrrrc}
    \toprule   
   \textbf{Nro } & \textbf{ OpenML name} & \textbf{OpenML did} & \textbf{ D } & \textbf{ N } & \textbf{ C } & \textbf{ nMaj } & \textbf{ nMin } & \textbf{ P} & \textbf{Meta} \\ 
  \midrule
  
  30 & iris &  61 &   5 & 150 &   3 &  50 &  50 & 1.00 & base \\ 
  31 & molecular-biology\_promoters & 164 &  59 & 106 &   2 &  53 &  53 & 1.00 & base \\ 
  32 & satimage & 182 &  37 & 6430 &   6 & 1531 & 625 & 0.41 & $\bullet$\\ 
  33 & baseball & 185 &  18 & 1340 &   3 & 1215 &  57 & 0.05 & $\bullet$\\ 
  34 & wine & 187 &  14 & 178 &   3 &  71 &  48 & 0.68 & base\\ 
  35 & eucalyptus & 188 &  20 & 736 &   5 & 214 & 105 & 0.49 & $\bullet$\\ 
  36 & Australian & 292 &  15 & 690 &   2 & 383 & 307 & 0.80 & $\bullet$\\ 
  37 & satellite\_image & 294 &  37 & 6435 &   6 & 871 & 275 & 0.32 & base \\ 
  38 & libras\_move & 299 &  91 & 360 &  11 &  24 &  11 & 0.46 & base \\ 
  39 & vowel & 307 &  13 & 990 &  11 &  90 &  90 & 1.00 & base \\
  40 & mammography & 310 &   7 & 11183 &   2 & 10923 & 260 & 0.02 & base \\ 
  41 & oil\_spill & 311 &  50 & 937 &   2 & 896 &  41 & 0.05 & $\bullet$\\ 
  42 & yeast\_ml8 & 316 & 117 & 2417 &   2 & 2383 &  34 & 0.01 & base \\ 
  43 & hayes-roth & 329 &   5 & 160 &   4 &  65 &   0 & 0.01 & base\\ 
  44 & monks-problems-1 & 333 &   7 & 556 &   2 & 278 & 278 & 1.00 & base\\ 
  45 & monks-problems-2 & 334 &   7 & 601 &   2 & 395 & 206 & 0.52 & base \\ 
  46 & monks-problems-3 & 335 &   7 & 554 &   2 & 288 & 266 & 0.92 & base \\ 
  47 & SPECT & 336 &  23 & 267 &   2 & 212 &  55 & 0.26 & $\bullet$\\ 
  48 & grub-damage & 338 &   9 & 155 &   4 &  49 &  19 & 0.39 & $\bullet$\\ 
  49 & synthetic\_control & 377 &  62 & 600 &   6 & 100 & 100 & 1.00 & $\bullet$\\ 
  50 & analcatdata\_boxing2 & 444 &   4 & 132 &   2 &  71 &  61 & 0.86 & base \\ 
  51 & analcatdata\_boxing1 & 448 &   4 & 120 &   2 &  78 &  42 & 0.54 & base\\ 
  52 & analcatdata\_lawsuit & 450 &   5 & 264 &   2 & 245 &  19 & 0.08 & base\\ 
  53 & irish & 451 &   6 & 500 &   2 & 278 & 222 & 0.80 & $\bullet$ \\ 
  54 & analcatdata\_broadwaymult & 452 &   8 & 285 &   7 & 118 &  21 & 0.18& $\bullet$ \\ 
  55 & cars & 455 &   9 & 406 &   3 & 254 &  73 & 0.29 & $\bullet$\\ 
  56 & analcatdata\_authorship & 458 &  71 & 841 &   4 & 317 &  55 & 0.17 & base\\ 
  57 & analcatdata\_creditscore & 461 &   7 & 100 &   2 &  73 &  27 & 0.37 & base \\ 
  58 & backache & 463 &  33 & 180 &   2 & 155 &  25 & 0.16 & base\\ 
  59 & prnn\_synth & 464 &   3 & 250 &   2 & 125 & 125 & 1.00& $\bullet$ \\ 
  60 & schizo & 466 &  15 & 340 &   2 & 177 & 163 & 0.92 & $\bullet$\\ 
  61 & analcatdata\_dmft & 469 &   5 & 797 &   6 & 155 & 123 & 0.79 & base \\ 
  62 & profb & 470 &  10 & 672 &   2 & 448 & 224 & 0.50& $\bullet$ \\ 
  63 & analcatdata\_germangss & 475 &   6 & 400 &   4 & 100 & 100 & 1.00& base \\ 
  64 & biomed & 481 &   9 & 209 &   2 & 134 &  75 & 0.56 & $\bullet$\\ 
  65 & rmftsa\_sleepdata & 679 &   3 & 1024 &   4 & 404 &  94 & 0.23 & $\bullet$\\ 
  66 & visualizing\_livestock & 685 &   3 & 130 &   5 &  26 &  26 & 1.00 & $\bullet$\\ 
  67 & diggle\_table\_a2 & 694 &   9 & 310 &   9 &  41 &  18 & 0.44 & $\bullet$\\ 
    \bottomrule
    \end{tabular}
\end{table*}


\begin{table*}[h!]
\centering
\footnotesize

\setlength{\tabcolsep}{3pt}
 \caption{(Multi-class) classification OpenML datasets (68 to 104) used in experiments. For each dataset it is shown: the OpenML dataset name and id, the number of attributes (D), the number of examples (N), the number of classes (C), the number of examples belonging to the majority and minority classes (nMaj, nMin), the proportion between them (P), and whether the dataset was added to the enrichment step for meta-learning.}
 \label{app:data3}
\begin{tabular}{clcrrrrrrc}
    \toprule   
   \textbf{Nro } & \textbf{ OpenML name} & \textbf{OpenML did} & \textbf{ D } & \textbf{ N } & \textbf{ C } & \textbf{ nMaj } & \textbf{ nMin } & \textbf{ P} & \textbf{Meta} \\ 
  \midrule
  
   68 & ada\_prior & 1037 &  15 & 4562 &   2 & 3430 & 1132 & 0.33 & $\bullet$\\ 
  69 & ada\_agnostic & 1043 &  49 & 4562 &   2 & 3430 & 1132 & 0.33 & $\bullet$\\ 
  70 & jEdit\_4.2\_4.3 & 1048 &   9 & 369 &   2 & 204 & 165 & 0.81 & $\bullet$\\ 
  71 & pc4 & 1049 &  38 & 1458 &   2 & 1280 & 178 & 0.14 & $\bullet$\\ 
  72 & pc3 & 1050 &  38 & 1563 &   2 & 1403 & 160 & 0.11 & $\bullet$\\ 
  73 & mc2 & 1054 &  40 & 161 &   2 & 109 &  52 & 0.48 & $\bullet$\\ 
  74 & mc1 & 1056 &  39 & 9466 &   2 & 9398 &  68 & 0.01 & $\bullet$\\ 
  75 & ar4 & 1061 &  30 & 107 &   2 &  87 &  20 & 0.23 & $\bullet$\\ 
  76 & kc2 & 1063 &  22 & 522 &   2 & 415 & 107 & 0.26 & $\bullet$\\ 
  77 & ar6 & 1064 &  30 & 101 &   2 &  86 &  15 & 0.17 & $\bullet$\\ 
  78 & kc3 & 1065 &  40 & 458 &   2 & 415 &  43 & 0.10 & $\bullet$\\ 
  79 & kc1-binary & 1066 &  95 & 145 &   2 &  85 &  60 & 0.71 & $\bullet$\\ 
  80 & kc1 & 1067 &  22 & 2109 &   2 & 1783 & 326 & 0.18 & $\bullet$\\ 
  81 & pc1 & 1068 &  22 & 1109 &   2 & 1032 &  77 & 0.07 & $\bullet$\\ 
  82 & pc2 & 1069 &  37 & 5589 &   2 & 5566 &  23 & 0.00 & $\bullet$\\ 
  83 & mw1 & 1071 &  38 & 403 &   2 & 372 &  31 & 0.08 & $\bullet$\\ 
  84 & jEdit\_4.0\_4.2 & 1073 &   9 & 274 &   2 & 140 & 134 & 0.96 & $\bullet$\\ 
  85 & datatrieve & 1075 &   9 & 130 &   2 & 119 &  11 & 0.09 & $\bullet$\\ 
  86 & PopularKids & 1100 &  11 & 478 &   3 & 247 &  90 & 0.36 & $\bullet$\\ 
  87 & teachingAssistant & 1115 &   7 & 151 &   3 &  52 &  49 & 0.94 & $\bullet$\\ 
  88 & musk & 1116 & 170 & 6598 &   2 & 5581 & 1017 & 0.18 & $\bullet$\\ 
  89 & badges2 & 1121 &  12 & 294 &   2 & 210 &  84 & 0.40 & $\bullet$\\ 
  90 & pc1\_req & 1167 &   9 & 320 &   2 & 213 & 107 & 0.50 & $\bullet$\\ 
  91 & MegaWatt1 & 1442 &  38 & 253 &   2 & 226 &  27 & 0.12 & $\bullet$\\ 
  92 & PizzaCutter1 & 1443 &  38 & 661 &   2 & 609 &  52 & 0.09 & $\bullet$\\ 
  93 & PizzaCutter3 & 1444 &  38 & 1043 &   2 & 916 & 127 & 0.14 & $\bullet$\\ 
  94 & CostaMadre1 & 1446 &  38 & 296 &   2 & 258 &  38 & 0.15 & $\bullet$\\ 
  95 & CastMetal1 & 1447 &  38 & 327 &   2 & 285 &  42 & 0.15 & $\bullet$\\ 
  96 & PieChart1 & 1451 &  38 & 705 &   2 & 644 &  61 & 0.09 & $\bullet$\\ 
  97 & PieChart2 & 1452 &  37 & 745 &   2 & 729 &  16 & 0.02 & $\bullet$\\ 
  98 & PieChart3 & 1453 &  38 & 1077 &   2 & 943 & 134 & 0.14& $\bullet$ \\ 
  99 & acute-inflammations & 1455 &   7 & 120 &   2 &  70 &  50 & 0.71 & base \\ 
  100 & appendicitis & 1456 &   8 & 106 &   2 &  85 &  21 & 0.25 & base\\ 
  101 & artificial-characters & 1459 &   8 & 10218 &  10 & 1416 & 600 & 0.42 & base \\ 
  102 & banknote-authentication & 1462 &   5 & 1372 &   2 & 762 & 610 & 0.80 & base \\ 
  103 & blogger & 1463 &   6 & 100 &   2 &  68 &  32 & 0.47 & $\bullet$\\ 
  104 & blood-transfusion-service-center & 1464 &   5 & 748 &   2 & 570 & 178 & 0.31 & base \\ 
   \bottomrule
    \end{tabular}
\end{table*}


\begin{table*}[h!]
\centering
\footnotesize

\setlength{\tabcolsep}{3pt}
 \caption{(Multi-class) classification OpenML datasets (105 to 141) used in experiments. For each dataset it is shown: the OpenML dataset name and id, the number of attributes (D), the number of examples (N), the number of classes (C), the number of examples belonging to the majority and minority classes (nMaj, nMin), the proportion between them (P), and whether the dataset was added to the enrichment step for meta-learning.}
 \label{app:data4}
\begin{tabular}{clcrrrrrrc}
    \toprule   
   \textbf{Nro } & \textbf{ OpenML name} & \textbf{OpenML did} & \textbf{ D } & \textbf{ N } & \textbf{ C } & \textbf{ nMaj } & \textbf{ nMin } & \textbf{ P} & \textbf{Meta} \\ 
  \midrule
  
  105 & breast-tissue & 1465 &  10 & 106 &   6 &  22 &  14 & 0.64 & base\\ 
  106 & cardiotography & 1466 & 35 & 2126 & 10 & 579 & 53 & 0.091 & base \\
  107 & cnae-9 & 1468 & 857 & 1080 &   9 & 120 & 120 & 1.00 & $\bullet$\\ 
  108 & fertility & 1473 &  10 & 100 &   2 &  88 &  12 & 0.14 & base \\ 
  109 & first-order-theorem-proving & 1475 &  52 & 6118 &   6 & 2554 & 486 & 0.19 & base \\ 
  110 & hill-valley & 1479 & 101 & 1212 &   2 & 606 & 606 & 1.00 & base \\ 
  111 & ilpd & 1480 &  11 & 583 &   2 & 416 & 167 & 0.40 & base \\ 
  112 & lsvt & 1484 & 311 & 126 &   2 &  84 &  42 & 0.50 & base \\ 
  113 & ozone-level-8hr & 1487 &  73 & 2534 &   2 & 2374 & 160 & 0.07 & base \\ 
  114 & parkinsons & 1488 &  23 & 195 &   2 & 147 &  48 & 0.33 & base \\ 
  115 & phoneme & 1489 &   6 & 5404 &   2 & 3818 & 1586 & 0.42 & base \\ 
  116 & one-hundred-plants-shape & 1492 &  65 & 1600 & 100 &  16 &  16 & 1.00 & base\\ 
  117 & qsar-biodeg & 1494 &  42 & 1055 &   2 & 699 & 356 & 0.51 & $\bullet$\\ 
  118 & qualitative-bankruptcy & 1495 &   7 & 250 &   2 & 143 & 107 & 0.75 & $\bullet$\\ 
  119 & ringnorm & 1496 &  21 & 7400 &   2 & 3736 & 3664 & 0.98 & $\bullet$\\ 
  120 & wall-robot-navigation & 1497 &  25 & 5456 &   4 & 2205 & 328 & 0.15 & base \\ 
  121 & sa-heart & 1498 &  10 & 462 &   2 & 302 & 160 & 0.53 & base \\ 
  122 & steel-plates-fault & 1504 &  34 & 1941 &   2 & 1268 & 673 & 0.53 & base\\ 
  123 & thoracic-surgery & 1506 &  17 & 470 &   2 & 400 &  70 & 0.18 & base \\ 
  124 & twonorm & 1507 &  21 & 7400 &   2 & 3703 & 3697 & 1.00 & $\bullet$\\ 
  125 & wdbc & 1510 &  31 & 569 &   2 & 357 & 212 & 0.59 & $\bullet$\\ 
  126 & wholesale-customers & 1511 &   9 & 440 &   2 & 298 & 142 & 0.48 & $\bullet$\\ 
  127 & heart-long-beach & 1512 &  14 & 200 &   5 &  56 &  10 & 0.18 & base \\ 
  128 & robot-failures-lp4 & 1519 &  91 & 117 &   3 &  72 &  21 & 0.29& $\bullet$ \\ 
  129 & robot-failures-lp5 & 1520 &  91 & 164 &   5 &  47 &  21 & 0.45 & $\bullet$\\ 
  130 & vertebra-column & 1523 &   7 & 310 &   3 & 150 &  60 & 0.40 & base \\ 
  131 & volcanoes-a2 & 1528 &   4 & 1623 &   5 & 1471 &  29 & 0.02 & $\bullet$\\ 
  132 & volcanoes-a3 & 1529 &   4 & 1521 &   5 & 1369 &  29 & 0.02 & $\bullet$\\ 
  133 & volcanoes-b1 & 1531 &   4 & 10176 &   5 & 9791 &  26 & 0.00 & $\bullet$\\ 
  134 & volcanoes-b3 & 1533 &   4 & 10386 &   5 & 10006 &  25 & 0.00 & $\bullet$\\ 
  135 & volcanoes-b5 & 1535 &   4 & 9989 &   5 & 9599 &  26 & 0.00 & $\bullet$\\ 
  136 & volcanoes-b6 & 1536 &   4 & 10130 &   5 & 9746 &  26 & 0.00 & $\bullet$\\ 
  137 & volcanoes-d2 & 1539 &   4 & 9172 &   5 & 8670 &  56 & 0.01 & $\bullet$\\ 
  138 & volcanoes-d3 & 1540 &   4 & 9285 &   5 & 8771 &  58 & 0.01& $\bullet$ \\ 
  139 & autoUniv-au1-1000 & 1547 &  21 & 1000 &   2 & 741 & 259 & 0.35 & base \\ 
  140 & autoUniv-au6-750 & 1549 &  41 & 750 &   8 & 165 &  57 & 0.35 & base \\
 141 & autoUniv-au6-400 & 1551 &  41 & 400 &   8 & 111 &  25 & 0.23 & base\\ 
     \bottomrule
    \end{tabular}
\end{table*}


\begin{table*}[h!]
\centering
\footnotesize

\setlength{\tabcolsep}{3pt}
 \caption{(Multi-class) classification OpenML datasets (142 to 182) used in experiments. For each dataset it is shown: the OpenML dataset name and id, the number of attributes (D), the number of examples (N), the number of classes (C), the number of examples belonging to the majority and minority classes (nMaj, nMin), the proportion between them (P), and whether the dataset was added to the enrichment step for meta-learning.}
 \label{app:data5}
\begin{tabular}{clcrrrrrrc}
    \toprule   
   \textbf{Nro } & \textbf{ OpenML name} & \textbf{OpenML did} & \textbf{ D } & \textbf{ N } & \textbf{ C } & \textbf{ nMaj } & \textbf{ nMin } & \textbf{ P} & \textbf{Meta} \\ 
  \midrule
  
  142 & autoUniv-au7-1100 & 1552 &  13 & 1100 &   5 & 305 & 153 & 0.50 & base\\ 
  143 & autoUniv-au7-500 & 1554 &  13 & 500 &   5 & 192 &  43 & 0.22 & base \\ 
  144 & acute-inflammations & 1556 &   7 & 120 &   2 &  61 &  59 & 0.97 & $\bullet$\\ 
  145 & bank-marketing & 1558 &  17 & 4521 &   2 & 4000 & 521 & 0.13 & base \\ 
  146 & breast-tissue & 1559 &  10 & 106 &   4 &  49 &  14 & 0.29 & $\bullet$\\ 
  147 & hill-valley & 1566 & 101 & 1212 &   2 & 612 & 600 & 0.98 & $\bullet$\\ 
  148 & wilt & 1570 &   6 & 4839 &   2 & 4578 & 261 & 0.06 & $\bullet$\\ 
  149 & SPECTF & 1600 &  45 & 267 &   2 & 212 &  55 & 0.26 & $\bullet$\\ 
  150 & PhishingWebsites & 4534 &  31 & 11055 &   2 & 6157 & 4898 & 0.80 \\ 
  151 & MiceProtein & 4550 &  82 & 1080 &   8 & 150 & 105 & 0.70& $\bullet$ \\ 
  152 & cylinder-bands & 6332 &  40 & 540 &   2 & 312 & 228 & 0.73& $\bullet$ \\ 
  153 & thyroid-allbp & 40474 &  27 & 2800 &   5 & 1632 &  31 & 0.02 & base \\ 
  154 & thyroid-allhyper & 40475 &  27 & 2800 &   5 & 1632 &  31 & 0.02 & base \\ 
  155 & LED-display-domain-7digit & 40496 &   8 & 500 &  10 &  57 &  37 & 0.65 & base \\ 
  156 & texture & 40499 &  41 & 5500 &  11 & 500 & 500 & 1.00 & base \\ 
  157 & cmc &  23 &  10 & 1473 &   3 & 629 & 333 & 0.53 & base\\ 
  158 & credit-a &  29 &  16 & 690 &   2 & 383 & 307 & 0.80 & $\bullet$\\ 
  159 & page-blocks &  30 &  11 & 5473 &   5 & 4913 &  28 & 0.01 & base \\ 
  160 & heart-h &  51 &  14 & 294 &   5 & 188 &   0 & 0.00 & base \\ 
  161 & banana & 1460 &   3 & 5300 &   2 & 2924 & 2376 & 0.81 & base \\ 
  162 & planning-relax & 1490 &  13 & 182 &   2 & 130 &  52 & 0.40 & $\bullet$\\ 
  163 & one-hundred-plants-margin & 1491 &  65 & 1600 & 100 &  16 &  16 & 1.00 & base \\ 
  164 & user-knowledge & 1508 &   6 & 403 &   5 & 129 &  24 & 0.19 & base  \\ 
  165 & volcanoes-a1 & 1527 &   4 & 3252 &   5 & 2952 &  58 & 0.02 & $\bullet$\\ 
  166 & volcanoes-a4 & 1530 &   4 & 1515 &   5 & 1365 &  29 & 0.02 & $\bullet$\\ 
  167 & autoUniv-au7-700 & 1553 &  13 & 700 &   3 & 245 & 214 & 0.87 & base\\ 
  168 & autoUniv-au6-1000 & 1555 &  41 & 1000 &   8 & 240 &  89 & 0.37 & base\\

  169 & live-disorders &   8 &   6 & 345 &  2 & 200 & 145 & 0.58 & base \\ 
  170 & autoUniv-au4-2500 & 1548 & 100 & 2500 &   3 & 1173 & 196  &  0.47 & base \\ 
  171 & cardiotocography v.2 (version 2) & 1560 &  35 & 2126 &  3 & 1655 & 176 & 0.78 & base \\ 
  172 & cloud & 210 &   6 & 108 &  4 & 32 & 10 &  0.30 & base \\ 
  173 & solar-flare & 173 &  12 & 1066 &  6 &  83 &  24 &  0.29 & base \\ 
  174 & heart-h v.3 (version 3) & 1565 &  13 & 294 &  5 & 188 & 15 &  0.64 & base \\ 
  175 & micro-mass & 1514 & 1300 & 360 &  10 & 36 & 36 & 1.00 & base \\ 
  176 & mfeat-factors &  12 & 217 & 2000 &  10 & 200 & 200 & 1.00 & base \\ 
  177 & mushroom &  24 &  21 & 8124 &   2 & 4208 & 3916 &  0.52 & base \\ 
  178 & nursery (v.3) & 1568 &   9 & 12958 &   4 & 4320 & 4044 &  0.93 & base \\ 
  179 & ozone\_level v.2 & 40735 &  72 & 2536 &   2 & 2460 & 76 &  0.97 & base \\
  180 & one-hundred-plants-texture & 1493 &  65 & 1600 & 100 & 16 & 16 &  1.00 & base \\ 
  181 & seeds & 1499 &   7 & 210 &   3 & 70 & 70 & 1.00 & base \\ 
  182 & semeion & 1501 & 257 & 1593 & 10 & 161 & 155 & 0.96 & base \\ 
 
 \bottomrule

\end{tabular}
\end{table*}


\newpage

\clearpage

\section{Hyperparameter distributions of the best solutions returned by the Irace tuning technique} 
\label{app:hpDistributions}


\begin{figure*}[h!]
    \centering
    \begin{tabular}{ccc}
        \subfloat[C]{\includegraphics[width = 1.5in]{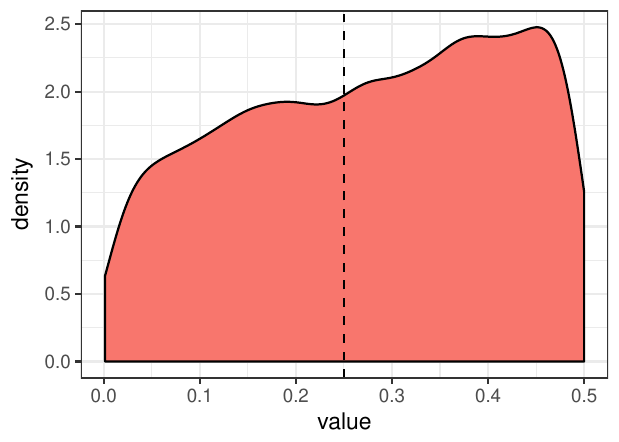}} &
        \subfloat[M]{\includegraphics[width = 1.5in]{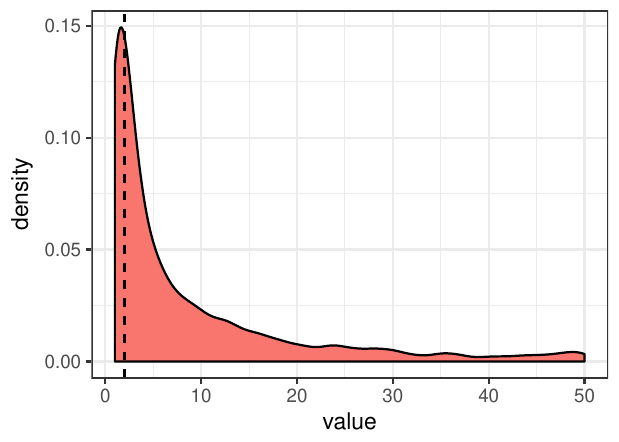}} &
        \subfloat[N]{\includegraphics[width = 1.5in]{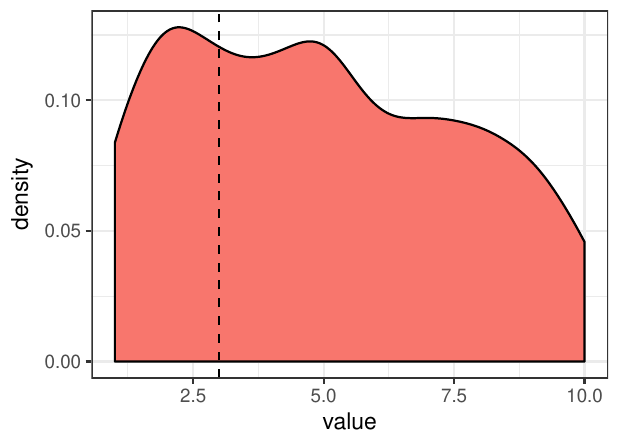}} \\
        \subfloat[R]{\includegraphics[width = 1.5in]{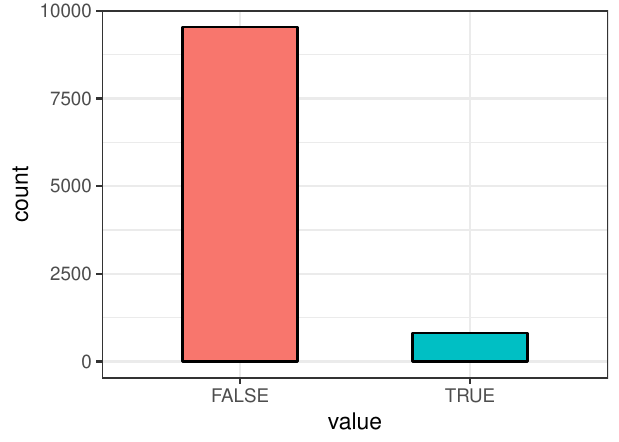}} & 
        \subfloat[O]{\includegraphics[width = 1.5in]{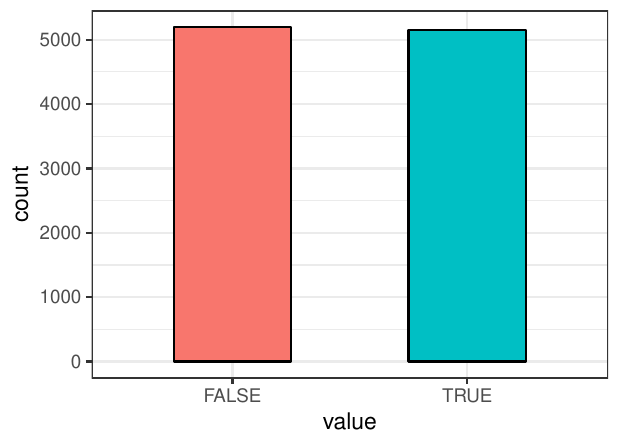}} & 
        \subfloat[B]{\includegraphics[width = 1.5in]{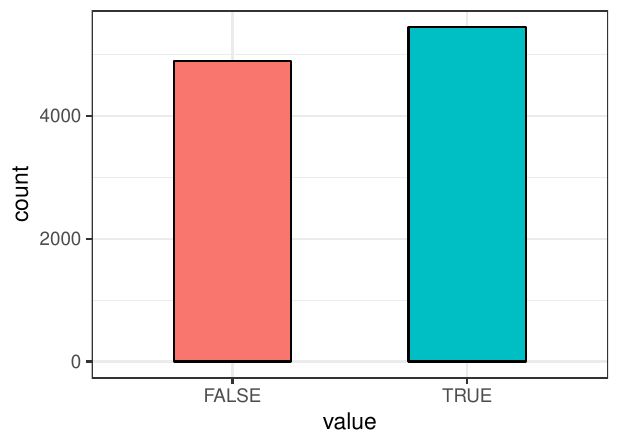}} \\ 
        \subfloat[A]{\includegraphics[width = 1.5in]{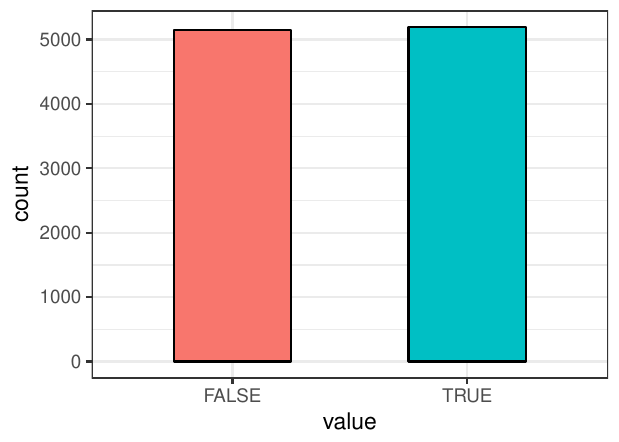}} &
        \subfloat[S]{\includegraphics[width = 1.5in]{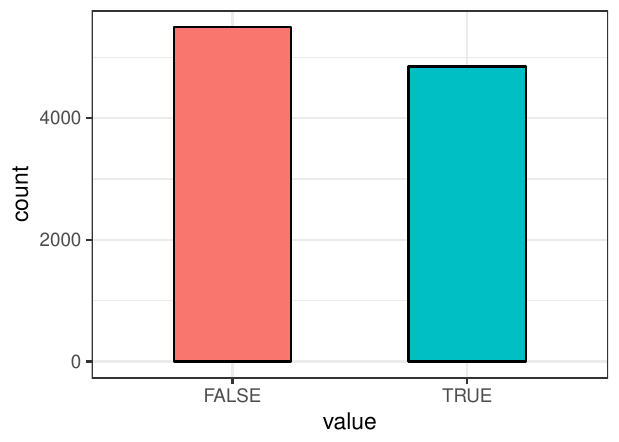}} & 
        \subfloat[J]{\includegraphics[width = 1.5in]{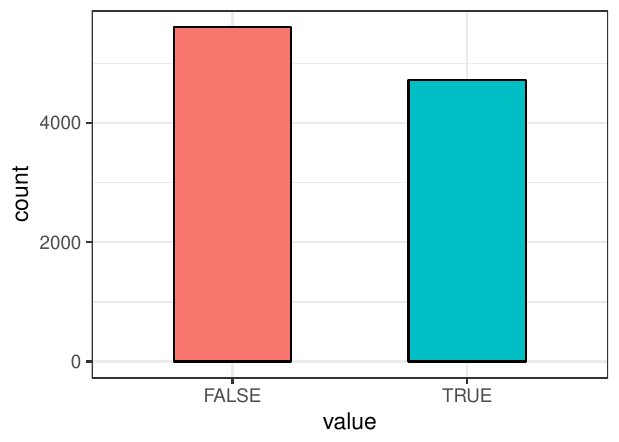}} 
    \end{tabular}
    \caption{Distribution of the J48 hyperparameters. Default values of the numerical hyperparameters are identified by black vertical dashed lines.}
    \label{fig:j48_distributions}
\end{figure*}


\begin{figure*}[th!]
    \centering
    \begin{tabular}{ccc}
        \subfloat[cp]{\includegraphics[width = 1.5in]{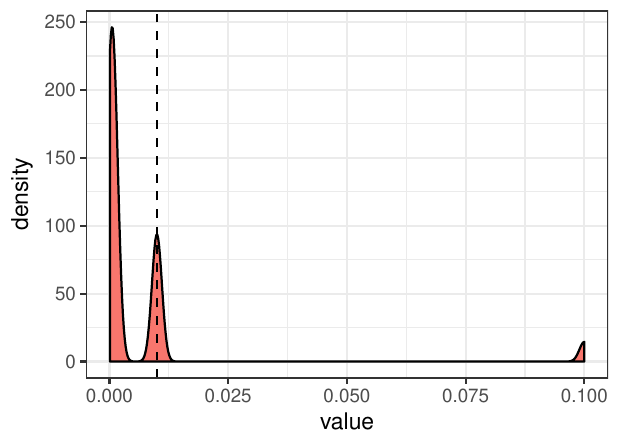}} &
        \subfloat[minbucket]{\includegraphics[width = 1.5in]{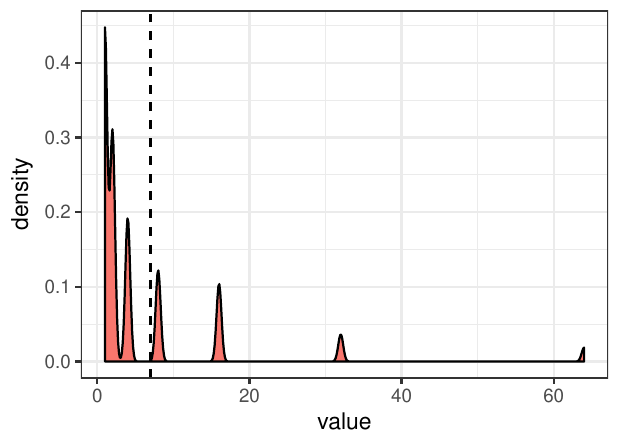}} &
        \subfloat[maxdepth]{\includegraphics[width = 1.5in]{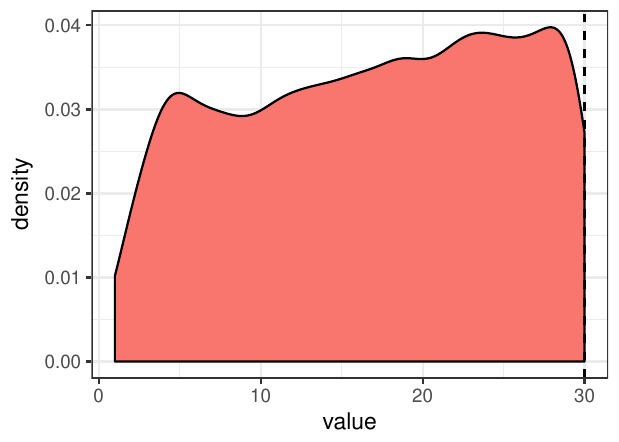}} \\
        \subfloat[minsplit]{\includegraphics[width = 1.5in]{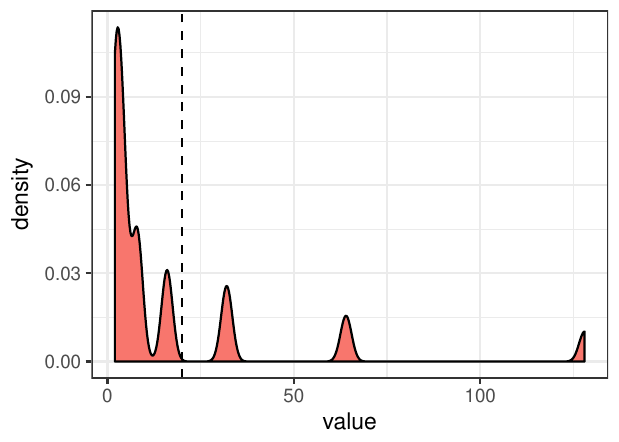}} & 
        \subfloat[usesurrogate]{\includegraphics[width = 1.5in]{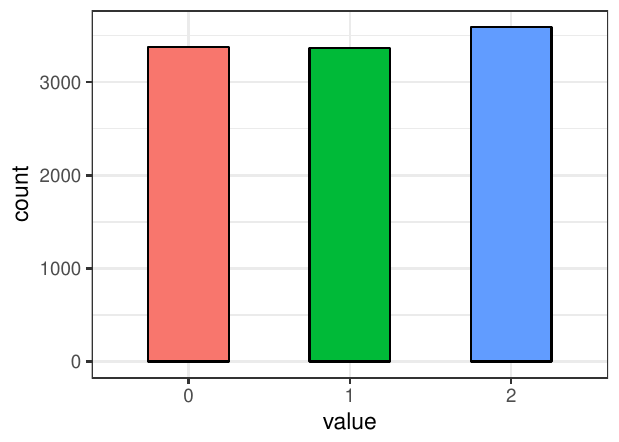}} & 
        \subfloat[surrogatestyle]{\includegraphics[width = 1.5in]{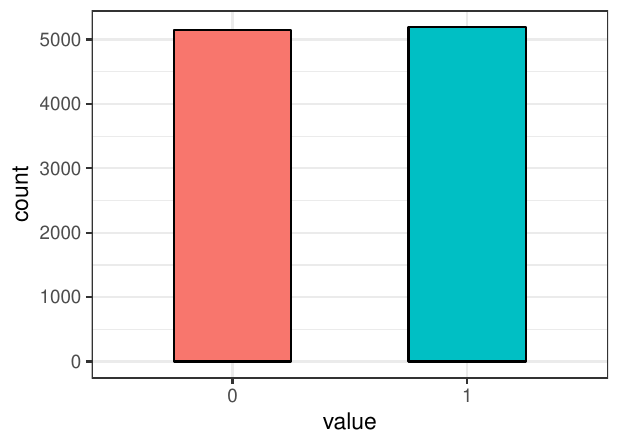}}
    \end{tabular}
    \caption{Distribution of the \acrshort{cart} hyperparameters. Default values of the numerical hyperparameters are identified by black vertical dashed lines.} 
    \label{fig:cart_distributions}
\end{figure*}


\end{appendices}

\end{document}